\documentclass[10pt,conference]{IEEEtran}

\ifCLASSINFOpdf
   \usepackage[pdftex]{graphicx}
   \DeclareGraphicsExtensions{.pdf,.jpeg,.png}
\else
   \usepackage[dvips]{graphicx}
   \DeclareGraphicsExtensions{.eps}
\fi

\usepackage{cite}
\usepackage{xcolor}
\usepackage{framed,multirow}
\usepackage{caption,subcaption}
\usepackage{booktabs}
\usepackage{url}
\usepackage{icomma}
\usepackage{array}
\usepackage{algorithmic}
\usepackage{hyphenat}
\usepackage{pgf,pgffor}
\usepackage[export]{adjustbox}
\usepackage[section]{placeins}
\usepackage{afterpage}

\usepackage[cmex10]{amsmath}
\usepackage{amsfonts,amsthm,amssymb,latexsym,mathtools}

\usepackage{fancyhdr}
\usepackage{tikz}
\usetikzlibrary{calc}

\begin{document}

\title{ConformalLayers: A non-linear sequential neural network with associative layers}

\author{\IEEEauthorblockN{Eduardo Vera Sousa, Leandro A. F. Fernandes, Cristina Nader Vasconcelos}
    \IEEEauthorblockA{Instituto de Computação, Universidade Federal Fluminense (UFF)\\
    Niterói, Rio de Janeiro, Brazil -- ZIP 24210--346\\
    Email: \{eduardovera, laffernandes, crisnv\}@ic.uff.br}
}

\maketitle

\begin{tikzpicture}[remember picture, overlay]%
\node at ($(current page.north) + (0,-0.5in)$) {Best Paper on Pattern Recognition and Related Field at SIBGRAPI 2021 -- 34th Conference on Graphics, Patterns and Images};%
\end{tikzpicture}%

\begin{abstract}
    Convolutional Neural Networks (CNNs) have been widely applied. But as the CNNs grow, the number of arithmetic operations and memory footprint also increase. Furthermore, typical non-linear activation functions do not allow associativity of the operations encoded by consecutive layers, preventing the simplification of intermediate steps by combining them. We present a new activation function that allows associativity between sequential layers of CNNs. Even though our activation function is non-linear, it can be represented by a sequence of linear operations in the conformal model for Euclidean geometry. In this domain, operations like, but not limited to, convolution, average pooling, and dropout remain linear. We take advantage of associativity to combine all the ``conformal layers'' and make the cost of inference constant regardless of the depth of the network.
\end{abstract}

\IEEEpeerreviewmaketitle

\section{Introduction}
\label{sec:introduction}

Convolutional Neural Networks (CNNs) are a type of neural network that have become popular in many areas with remarkable accuracy, including Computer Vision. As the network gets, however, higher is the number of operations~to perform, and more memory is required for both training and inference.

To make inference, a cheaper operation has become an active topic of research. The primary strategies for accomplishing this are compressing the networks~\cite{denton2014exploiting,hubara2016binarized,hubara2017quantized,frankle2018lottery,zhou2019deconstructing,pmlr-v130-omidvar21a}, or exploring data organization to fit hardware specificities~\cite{7783725,howard2017mobilenets,tan2019efficientnet,NEURIPS2020_ebd9629f}. But the former strategy leads to modifications to the original architecture. The latter depends on the hardware.

An interesting strategy for simplifying sequential layers of the CNNs would be the combination of adjacent layers by associativity. This idea would not change the network's architecture and, in the limit, all layers would be combined into a single operator to be applied to the input data. In this paper, the term \emph{associativity} refer to the mathematical property of a binary operation $\circ$ on a set $\mathcal{S}$ to satisfy the associative~law:
\begin{equation}
    \label{eq:associative_law}
    \left(x \circ y\right) \circ z = x \circ \left(y \circ z\right)\text{ for all } x, y, z \in \mathcal{S}\text{.}
\end{equation}

The idea of exploring operations' associativity to reduce computational costs has been widely explored in other areas, like computer graphics, for example. Unfortunately, composing layers considering existing artificial neural networks is only applicable to sequences of associative operations like convolutions and averages, since typical activation functions turn the layers of sequential networks non-associative.

This work presents a non-linear and differentiable activation function called ReSPro, which stands for Spherical \underline{Re}flection, \underline{S}caling, and \underline{Pro}jection. By representing the input data and any map of features produced by the network as points encoded as vectors in the conformal model for Euclidean geometry, ReSPro turns into a sequence of linear transformations in the conformal domain. By adding only one more coefficient to the input data, some linear transformations behave like non-linear, and become suitable as activation function in CNNs. ReSPro and typical operations like convolution, average pooling, dropout, flattening, padding, dilation, and stride can be encoded as rank-2 or rank-3 tensors whose product satisfies the associative law~(\ref{eq:associative_law}). We call ConformalLayers the conformal embedding of sequential layers of CNNs comprised of those operations. By associativity, one sparse matrix and one sparse rank-3 tensor computed after training are enough to encode a sequential CNN made up of any number of conformal layers, making the cost of inference remains constant.

The reduced number of operations required for operating the matrix and the rank-3 tensor resulting from the ConformalLayers with the input data makes our approach well-suited for processing a large number of images, even in devices with limited storage and computational capabilities. To fully explore the advantages of ReSPro and the ConformalLayers, we built a library on top of PyTorch. Such a framework allows the associativity between the sequential layers of CNNs and covers both training and inference processes.

\begin{figure}[!t]
    \centering
    \includegraphics[width=\columnwidth]{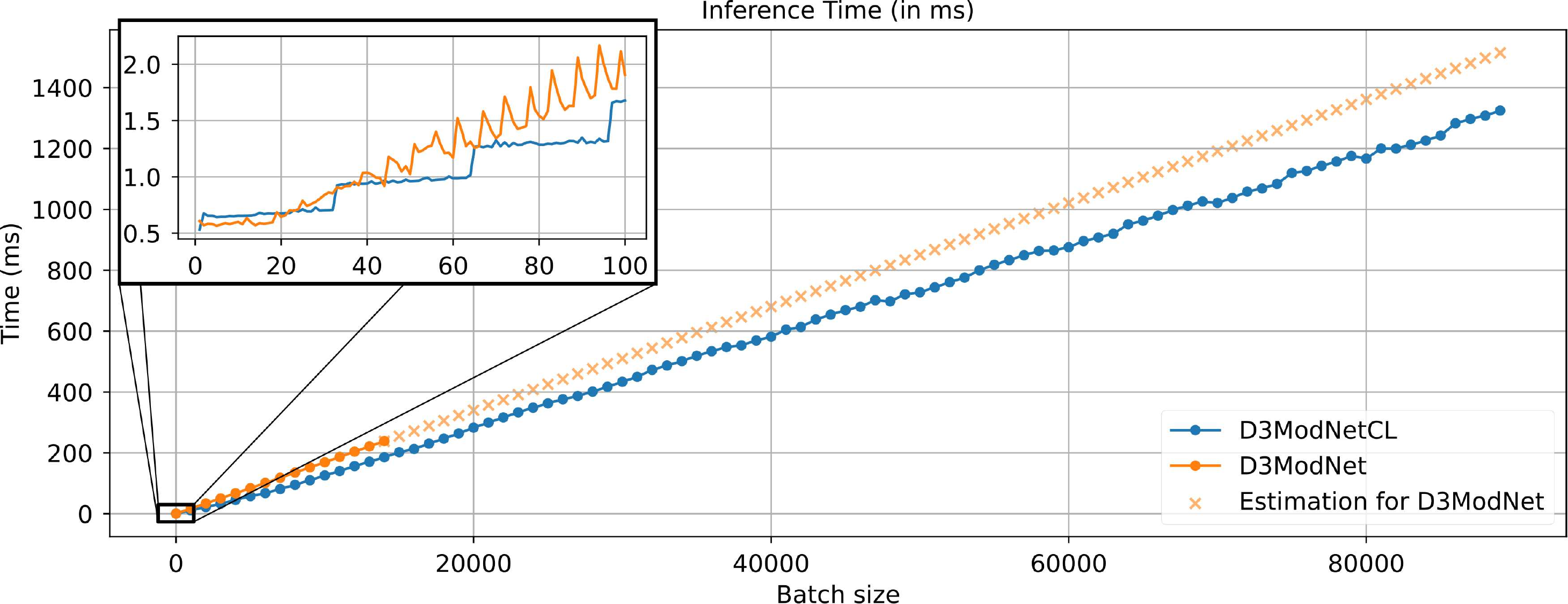}
    \caption{Inference times for the \texttt{D3ModNetCL}, a CNN whose feature extraction is implemented with ConformalLayers, and the \texttt{D3ModNet}, its counterpart with non-associative layers.}
    \label{fig:inference_D3NetCL_D3Net}
\end{figure}

Fig.~\ref{fig:inference_D3NetCL_D3Net} depicts the differences of inference time of the traditional implementation of a non-associative CNN (\texttt{D3ModNet}) and a similar network implemented using ConformalLayers (\texttt{D3ModNetCL}). The non-associative approach is limited to process 14K images simultaneously on an NVIDIA GTX 1050 Ti (orange line with circles). The ConformalLayers, on the other hand, was able to process 89K images simultaneously on the same hardware (blue line with circles). The orange crosses beyond the limitations of \texttt{D3ModNet} extrapolates its memory and execution times capabilities just for comparison.

Our main contributions can be summarized as:
\begin{itemize}
    \item A new non-linear and differentiable activation function;
    \item The first approach for sequential layers of CNNs where the linear and non-linear layers are associative;
    \item The analysis of the accuracy of conventional CNNs against similar CNNs with ConformalLayers; and
    \item The analysis of the computational performance of conventional CNNs against similar CNNs with ConformalLayers considering the number of layers and batch size.
\end{itemize}

    
\section{Related Work}
\label{sec:related_work}

We group the related works into activation functions, neural network compression, and computationally-efficient CNNs.

\paragraph*{Activation Functions}

vanishing gradient problem.
The sigmoid and hyperbolic tangent are S-shaped activation functions~\cite{GoodfellowEtAl2016}. While the sigmoid may lead to time convergence issues, the hyperbolic tangent mitigates this problem. ReLU~\cite{hahnloser2000digital} provides three main advantages over the hyperbolic tangent: eliminate the problem of vanishing gradient (common in S-shaped functions), add some sparsity to the data, and is computationally simple, although it can map many inputs to zero, preventing the network from learning. GeLU~\cite{hendrycks2016gaussian} is a non-monotonic non-linear activation function that weights the input by the standard Gaussian cumulative distribution function. Although it is slightly costly, it can provide similar or better results than ReLU. Swish~\cite{ramachandran2017swish} is another non-monotonic activation function that outperforms ReLU-like functions, but with higher computational cost.

To the best of our knowledge, all non-linear activation functions described in the literature are non-associative with linear operations typically used in CNNs. Therefore, they prevent the combination of sequential layers through associativity.

\paragraph*{Neural Network Compression}

Denton~\emph{et al.}~\cite{denton2014exploiting} post-process the trained neural networks to iteratively compress each layer and then fixing the accuracy, which presented a $2\times$ speedup at the cost of $1\%$ of accuracy. 

Hubara \emph{et al.}~\cite{hubara2016binarized} presented a network model using binarization, allowing binary operations instead of products during the inference by the cost of a small percent of accuracy~\cite{hubara2017quantized}.

The lottery ticket hypothesis was postulated by Frankle and Carbin~\cite{frankle2018lottery}. They state that training a neural network with random weights will most likely provide good results. Still, probably there is a sub-network that is trainable with lower cost and fewer parameters, while providing similar results. Zhou~\emph{et al.}~\cite{zhou2019deconstructing} used masking criteria to assess which weights should be pruned in this approach.

By partitioning the set of convolutional filters and using these sets as inputs for an auxiliary neural network, Omidvar~\emph{et al.}~\cite{pmlr-v130-omidvar21a} generated reusable filters, reducing the number of network parameters. Tetko~\cite{tetko2002associative} presented an ensemble of neural networks and $k$-NN which uses the distance between the predicted data and the ground truth to improve the prediction by adjusting the bias on the networks. Both approaches are called ``associative'' but its important to emphasize that associativity in this paper is related to the property in~(\ref{eq:associative_law}).

The neural network compression techniques discussed here seek to find different networks from those proposed initially or assume less precise data types. We state that it is possible to improve computational performance by combining layers of the network without changing its architecture, as long as an activation function that allows associativity is adopted.

\paragraph*{Computationally-Efficient CNNs}

YOLO~\cite{redmon2016you} and SqueezeDet~\cite{wu2017squeezedet} are popular real-time CNNs for object detection and classification. While the former is for general purpose, the latter is tailored to autonomous driving.

Alwani~\emph{et al.}~\cite{7783725} leveraged caching aspects in hardware accelerators by observing that each point of a hidden layer depends on a specific region of the input.

MobileNets~\cite{howard2017mobilenets} and EfficientNets~\cite{tan2019efficientnet} are approaches that change the size of a model to fit a resource budget. While the first rely on model depth and image resolution to handle devices with limited capacity, the latter proposes a compound parameter to scale the neural network. In~\cite{NEURIPS2020_ebd9629f}, a RNN-based pooling layer is presented to perform data down-sampling as a solution to~RAM and energy problems in edge devices.


\section{ConformalLayers}
\label{sec:conformal_layers}

In this section, we deep dive into the formulation and implementation of the ReSPro function and the ConformalLayers.

\subsection{ReSPro}
\label{sec:respro}

\begin{figure}[!t]
    \centering
    \begin{subfigure}[b]{\columnwidth}
        \centering
        \includegraphics[width=0.65\textwidth]{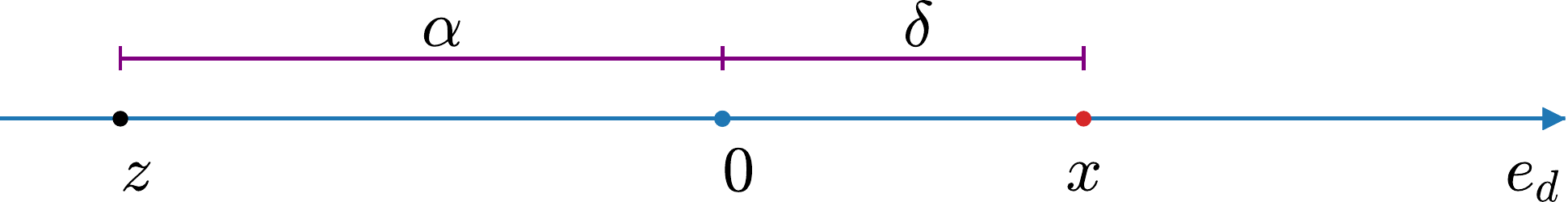}
        \caption{Data is encoded as points in a $d$-dimensional Cartesian space. Here, \mbox{$d = 1$} and the input data are points $z$ and $x$.}
        \label{fig:step1}
    \end{subfigure}
    \begin{subfigure}[b]{\columnwidth}
        \centering
        \includegraphics[width=0.65\textwidth]{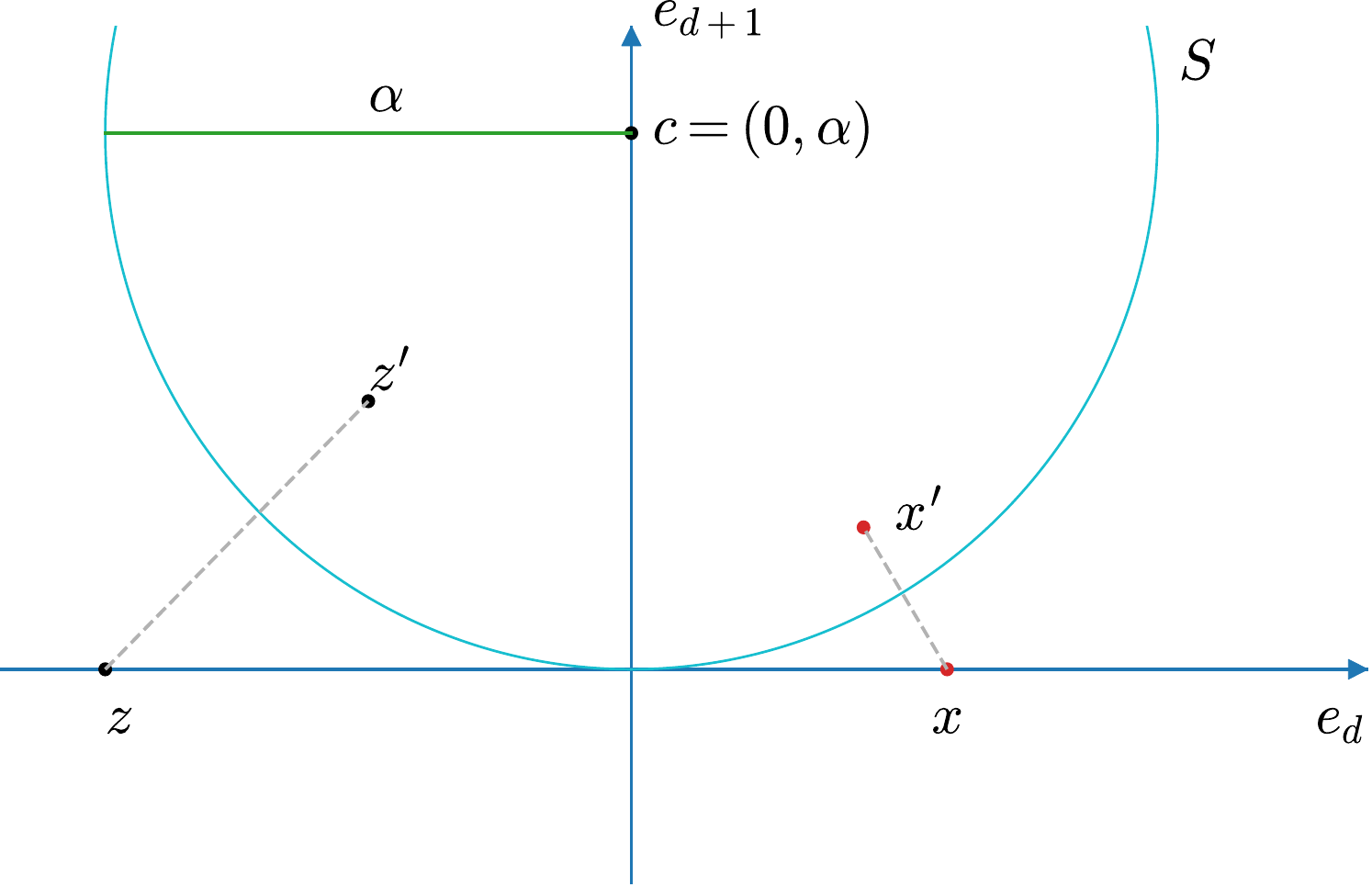}
        \caption{Spherical reflection on the sphere $S$ with center $c$ and radius $\alpha$ produces $z'$ and $x'$. The dimension $e_{d+1}$ was added to the Cartesian space to make $S$ tangent to the original $\mathbb{R}^d$ space.}
        \label{fig:step2}
    \end{subfigure}
    \begin{subfigure}[b]{\columnwidth}
        \centering
        \includegraphics[width=0.65\textwidth]{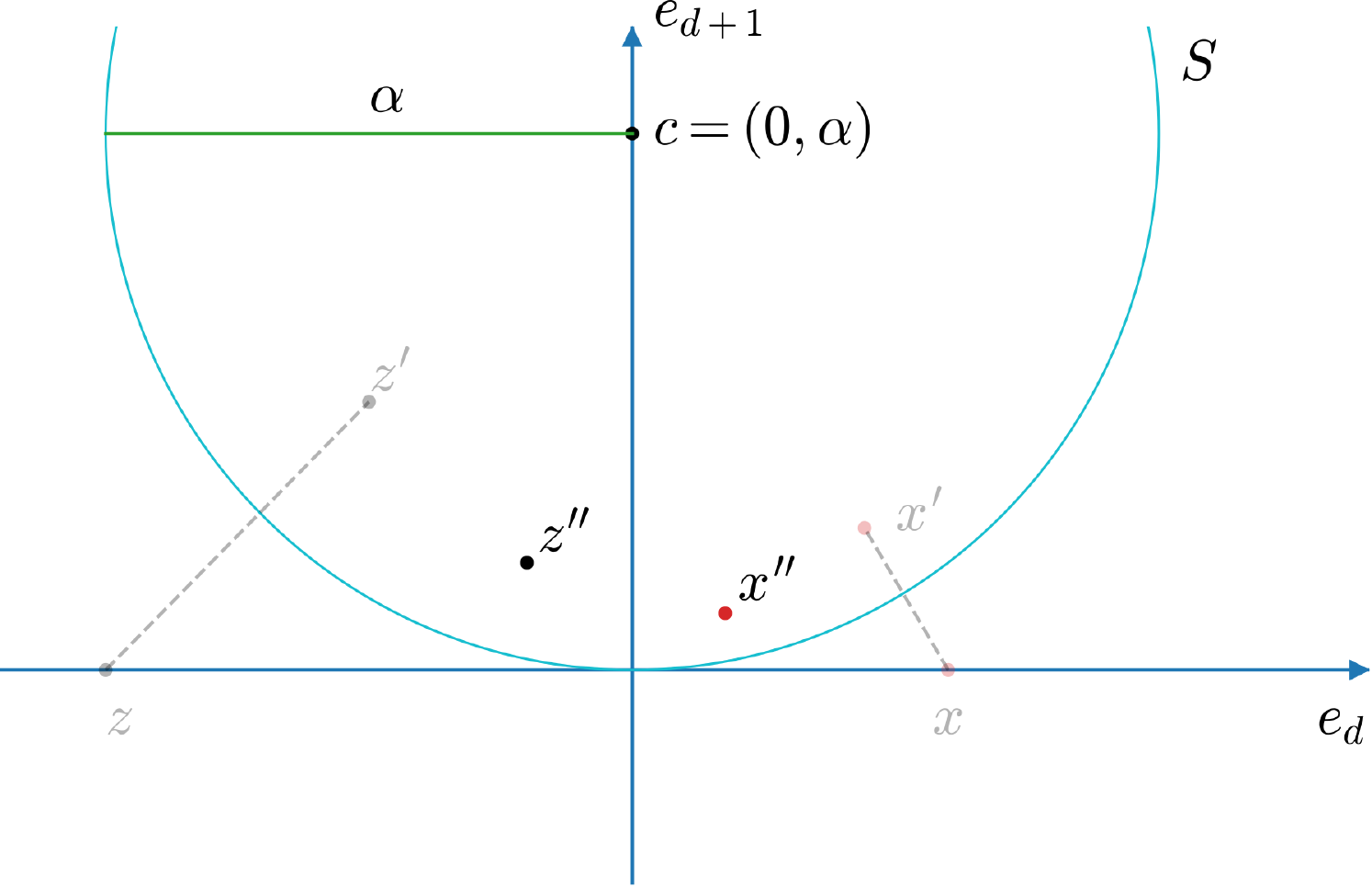}
        \caption{Using isotropic scaling, points $z'$ and $x'$ are mapped to $z''$ and~$x''$.}
        \label{fig:step3}
    \end{subfigure}
    \begin{subfigure}[b]{\columnwidth}
        \centering
        \includegraphics[width=0.65\textwidth]{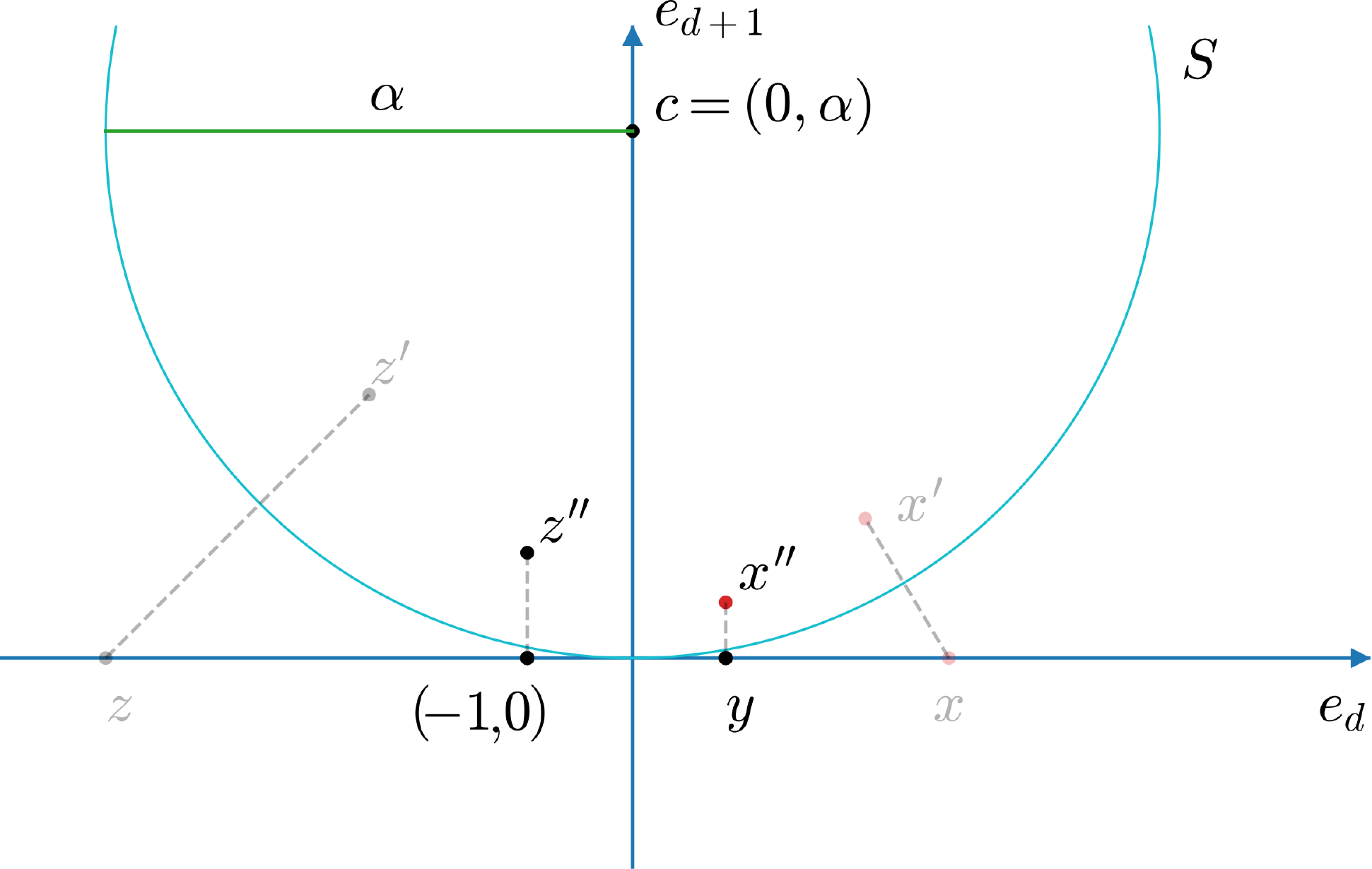}
        \caption{The final projection maps $z''$ and $x''$ to the original $\mathbb{R}^d$ space.}
        \label{fig:step4}
    \end{subfigure}
    \caption{Mapping performed by the ReSPro function.}
    \label{fig:step-by-step}
\end{figure}

Without loss of generality, we consider that the input and output data of the function are finite points in a multidimensional Cartesian space. Each coordinate of these points corresponds to a coefficient of the data. For example, a feature map with \mbox{$10 \times 10$} pixels and $3$ channels is a point \mbox{$x = \left(x_1, x_2, \cdots, x_d\right)$} with \mbox{$d = 10 \times 10 \times 3 = 300$} coordinates, which after being transformed by ReSPro maps to \mbox{$y = \left(y_1, y_2, \cdots, y_d\right)$}. The $x$-to-$y$ mapping involves one non-linear and two linear transformations applied to $x$. Fig.~\ref{fig:step-by-step} illustrates the step-by-step mapping performed by ReSPro. For sake of simplicity, we assume \mbox{$d = 1$} in this example.

Let $z$ and $x$ be two points lying on the $e_d$ axis (Fig.~\ref{fig:step1}), distant, respectively, $\alpha$ and $\delta$ units from the origin, for \mbox{$0 \leq \delta \leq \alpha$}. In Fig.~\ref{fig:step2}, we deliberately added the extra dimension $e_{d+1}$ to the Cartesian space and placed the hypersphere $S$ (a circle, in this example) with radius $\alpha$ and center \mbox{$c = \left(0, \alpha\right)$}. In this new space, \mbox{$z = \left(z_d, z_{d+1}\right) = \left(-\alpha, 0\right)$} and \mbox{$x = \left(x_d, x_{d+1}\right) = \left(\delta, 0\right)$}. The addition of the dimension $e_{d+1}$ to the Cartesian coordinate system is key for defining the non-linear transformation in ReSPro as the spherical reflection of points on $S$. In Fig.~\ref{fig:step2}, the spherical reflection maps $z$ and $x$ to, respectively, $z'$ and~$x'$.

Spherical reflection causes points outside the hyperspherical mirror to produce images inside the mirror. The distance of the imaged point to the center of the hypersphere decreases non-linearly as the distance of the original point to the center increases. Notice in Fig.~\ref{fig:step2} that the distance between points $z'$ and $c$ is smaller than the distance between $x'$ and $c$ since $z$ is more distant from $c$ than $x$. At the limit, points at infinity map to $c$, while points on the hypersphere remain unchanged. By construction, we do not have to care about the reflection of points inside the hyperspherical mirror because $S$ is tangent to the space spanned by \mbox{$\{e_1, e_2, \cdots, e_d\}$}, and all input points lie in this space (\emph{i.e.,}~the coordinate for $e_{d+1}$ is zero).

Two linear transformations are applied after spherical reflection. The first is isotropic scaling by a factor of \mbox{$2 / \alpha$}, illustrated in Fig.~\ref{fig:step3}, which maps $z'$ and $x'$ to $z''$ and $x''$, respectively. The second transformation is the orthogonal projection to the original $d$-dimensional Cartesian space. As can be seen in Fig.~\ref{fig:step4}, $z''$ projects to \mbox{$(-1, 0)$} while $x''$ projects to \mbox{$y = \left(y_d, 0\right)$}. The reasoning for performing scaling followed by projection is to map points that are $\alpha$ units away from the origin of the Cartesian space (\emph{e.g.,}~$z$) to points $1$ unit away from the origin. The projection also makes the coordinate associated with the extra dimension $e_{d+1}$ equal to zero, which can be removed from the resulting point since it is constant.

Formally, ReSPro is a mapping
\begin{equation}
    \label{eq:respro_mapping}
    f : x \rightarrow y\text{, subject to }\lVert x\rVert_2 \in \left[0, \alpha\right]\text{ and }\lVert y\rVert_2 \in \left[0, 1\right]\text{,}
\end{equation}
where \mbox{$x, y \in \mathbb{R}^{d}$} are, respectively, the input and output data represented as points, and \mbox{$\lVert p\rVert_2$} denotes the $L^2$-norm of $p$.

\begin{figure}[!t]
    \centering
    \begin{subfigure}[b]{0.47\columnwidth}
         \centering
         \includegraphics[width=\textwidth]{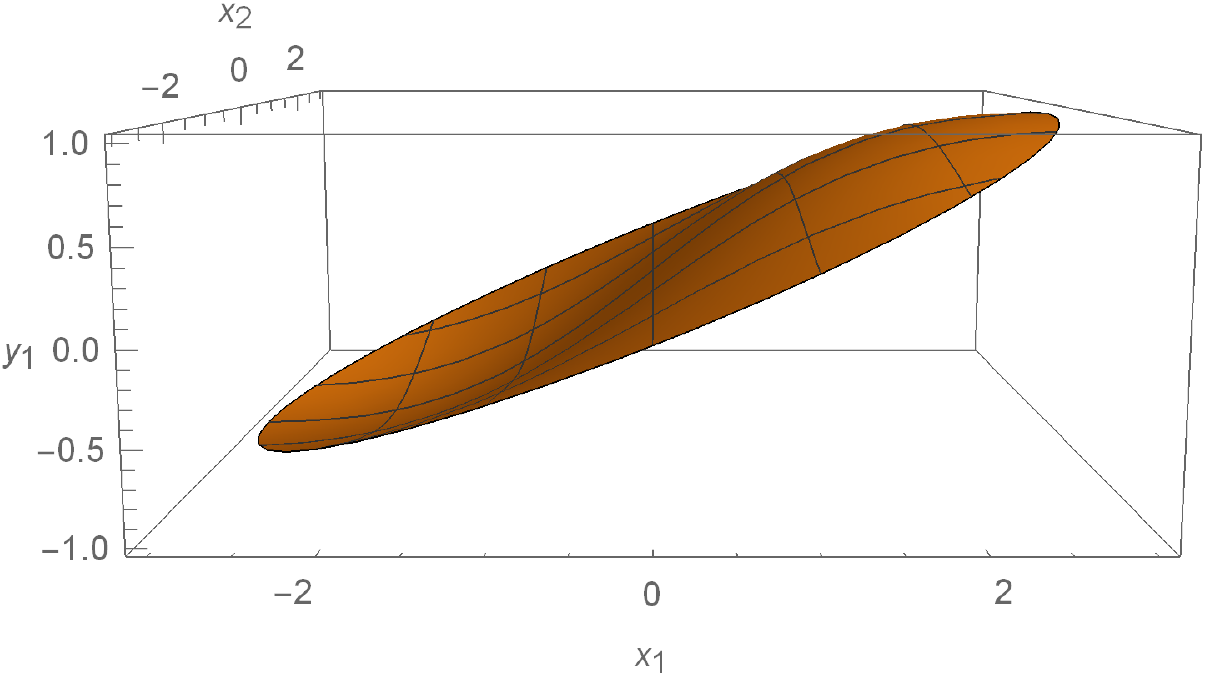}
         \label{fig:curve_x}
     \end{subfigure}
     \begin{subfigure}[b]{0.47\columnwidth}
         \centering
         \includegraphics[width=\textwidth]{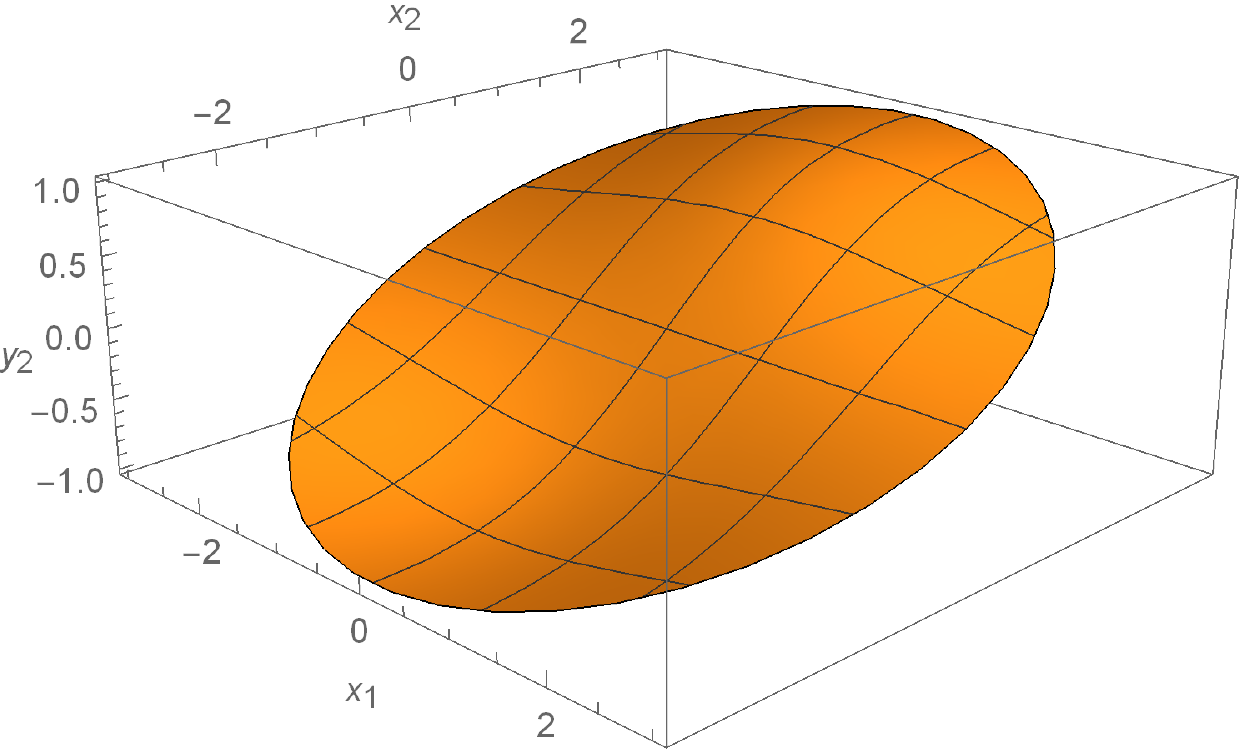}
         \label{fig:curve_y}
     \end{subfigure}
    \caption{The non-linear mapping produced by ReSPro. The surfaces in (a) and (b) show, respectively, the coordinates $y_1$ and $y_2$ of points $y$ resulting from applying ReSPro to points \mbox{$x = \left(x_1, x_2\right)$}. Here, we assume \mbox{$\alpha = 3$}, \emph{i.e.,}~\mbox{$\lVert x\rVert_2 \leq 3$}.}
    \label{fig:respro}
\end{figure}

Fig.~\ref{fig:respro} depicts the output of ReSPro for points \mbox{$x = \left(x_1, x_2\right)$} in $\mathbb{R}^2$ with \mbox{$\alpha = 3$}. Notice that the disc shape is due to the region where the function is defined, \emph{i.e.,}~\mbox{$\lVert x\rVert_2 \in \left[0, \alpha\right]$}.

ReSPro is non-linear, differentiable, invertible, and associative to linear function typically used in CNNs. The verification of the first three properties is straightforward. We demonstrate the last property in Section~\ref{sec:conformal_layers_tensors}. In addition, ReSPro is a global activation function. In practical terms, it means that, differently from other activation functions which perform per element transformation,
ReSPro activates all the elements simultaneously. This is equivalent to have element-wise activation followed by the normalization to ensure \mbox{$\lVert y\rVert_2 \in \left[0, 1\right]$}.

\subsection{Tensor Representation of ReSPro}
\label{sec:respro_tensors}

We use the geometric algebra of the conformal model for Euclidean geometry~\cite{dorst2010geometric} to perform spherical reflection as orthogonal transformation (hence, a linear transformation) and combine it to isotropic scaling and orthogonal projection. The representational space of the conformal model has two extra dimensions and a degenerate metric~\cite{dorst2010geometric}. The \mbox{$(d + 1)$}-dimensional Cartesian space used in Section~\ref{sec:respro} is embedded in a \mbox{$(d + 3)$}-dimensional space with basis vectors \mbox{$\{e_1, e_2, \cdots, e_{d+1}, e_o, e_\infty\}$}. The extra dimensions $e_o$ and $e_\infty$ are geometrically interpretable as the point at the origin and the point at infinity, respectively. In the conformal model, a finite point with coordinates \mbox{$p = \left(p_1, p_2, \cdots, p_d, p_{d+1}\right)$} is encoded by the $(d + 3)$-dimensional vector:
\begin{equation}
    \label{eq:finite_point}
    P = \gamma \left(p_1, p_2, \cdots, p_d, p_{d+1}, 1, \frac{1}{2} \sum_{i=1}^{d+1} (p_{i})^{2}\right)^\intercal\text{,}
\end{equation}
where ${}^\intercal$~denotes matrix transposition and the scalar \mbox{$\gamma \neq 0$} does not change the practical interpretation of $P$ as the point $p$.

The algebraic manipulation that takes the ReSPro function from its formulation in geometric algebra to tensor algebra is quite involved. Due to space restrictions, it is presented in the Supplementary Material. Here, it is sufficient to show that a \mbox{$d$-dimensional} point \mbox{$x = \left(x_1, x_2, \cdots, x_{d}\right)$} encoding input data will be represented by the $(d + 1)$-dimensional vector:
\begin{equation}
    \label{eq:input_vector}
    X = \left(x'_1, x'_2, \cdots, x'_d, x'_o\right)^\intercal\text{,}
\end{equation}
such that \mbox{$x_i = x'_i / x'_o$} and \mbox{$x'_o \neq 0$}, for \mbox{$i \in \{1, 2, \cdots, d\}$}. Vector $X$ is mapped to $Y$ by the ReSPro function as:
\begin{align}
    \label{eq:respro_without_tensors}
    Y &= \left(y'_1, y'_2, \cdots, y'_d, y'_o\right)^\intercal \nonumber \\
    &= \left(x'_1, x'_2, \cdots, x'_d, \frac{\alpha}{2} x'_{o} + \frac{1}{2 \alpha} \sum_{i=1}^{d} (x'_{i})^{2}\right)^\intercal\text{,}
\end{align}
such that \mbox{$y_i = y'_i / y'_o$} and \mbox{$y'_o \neq 0$}, for \mbox{$i \in \{1, 2, \cdots, d\}$}.

In~(\ref{eq:input_vector}) and~(\ref{eq:respro_without_tensors}), $X$ and $Y$ are simplified version of $P$~(\ref{eq:finite_point}). They do not include the coefficient related to $e_{d+1}$, since this coefficient is always zero (see Section~\ref{sec:respro}), nor the coefficient related to $e_\infty$, because it can be computed from the other coefficient. The $x'_o$ and $y'_o$ coefficients are related to the basis vector $e_o$ and act as homogeneous coordinates.

The tensor representation of ReSPro applied to $X$ is obtained by rewritten~(\ref{eq:respro_without_tensors})~as:
\begin{equation}
    \label{eq:respro}
    Y = \left(y'_1, y'_2, \cdots, y'_d, y'_o\right)^\intercal = \left(F_M + F_T X\right) X\text{,}
\end{equation}
where
\begin{equation*}
    F_M = \begin{pmatrix}
        1 & \cdots & 0 & 0 \\ 
        \vdots & \ddots & \vdots & \vdots \\ 
        0 & \cdots & 1 & 0 \\ 
        0 & \cdots & 0 & \frac{\alpha}{2} \\ 
    \end{pmatrix}
    \text{ and }
    F_T X = \begin{pmatrix}
        0 & \cdots & 0 & 0 \\ 
        \vdots & \ddots & \vdots & \vdots \\ 
        0 & \cdots & 0 & 0 \\ 
        \frac{x'_1}{2 \alpha} & \cdots & \frac{x'_d}{2 \alpha} & 0 \\ 
    \end{pmatrix}\text{.}
\end{equation*}
$F_M$ is a \mbox{$(d + 1) \times (d + 1)$} diagonal matrix, and $F_T$ is a rank-3 tensor of size \mbox{$(d+1) \times (d+1) \times (d+1)$} filled with zeros, except for the slice at the bottom, which is the diagonal matrix:
\begin{equation*}
    F_{T \left[d + 1\right]} = \begin{pmatrix}
        \frac{1}{2 \alpha} & \cdots & 0 & 0 \\ 
        \vdots & \ddots & \vdots & \vdots \\ 
        0 & \cdots & \frac{1}{2 \alpha} & 0 \\ 
        0 & \cdots & 0 & 0 \\ 
    \end{pmatrix}\text{.}
\end{equation*}

\subsection{Tensor Representation of ConformalLayers}
\label{sec:conformal_layers_tensors}

Recall that $X$ in~(\ref{eq:respro}) represents a finite point \mbox{$x \in \mathbb{R}^{d}$}, whose coordinates are the coefficients of some feature map produced by some CNN layer. It is well known that linear function typically used in layers and their configurations can be written in matrix form and applied to vectors by matrix multiplication. For instance, Toeplitz matrices encode $n$-dimensional discrete-time convolutions and can be modified to encode valid cross-correlation~\cite{GoodfellowEtAl2016}; average pooling is the mean filter~\cite{GonzalezWoods2008}, a particular case of convolution with constant weights; and configurations such as padding, dilation, and stride can be encoded by matrices composed of zeros and ones (see the Supplementary Material). By writing these and other operations in the matrix form, the associativity of the matrix product allows the composition of operations to produce matrices $U$ which, when multiplied by a vector $X$, produce the vector \mbox{$Z = U X$} representing the output of linear layers. By replacing $X$ in~(\ref{eq:respro}) by $Z$ and combining $U$ with $F_M$ and $F_T$, we write:
\begin{equation}
    \label{eq:layer_function}
    \begin{split}
        Y = \mathcal{L}(X) &= \left(F_M + F_T \left(U X\right)\right) \left(U X\right) \\
            &= \left(F_M U + (U^\intercal F_T^\intercal U)^\intercal X\right) X\text{,}
    \end{split}
\end{equation}
where $\mathcal{L}$ is \emph{the conformal layer function}, a sequence of linear operations applied to $X$, followed by the application of the ReSPro activation function. Here, ${}^\intercal$~denotes the transposition of the first two dimensions of tensors and matrix transposition.

In CNN architectures, it is typical for linear layers and activation functions to be interspersed. A sequence of $k$ conformal layers applied to $X$ can be written as:
\begin{equation}
    \label{eq:recursive_function}
    Y^{(k)} = \mathcal{L}^{(k)}(\mathcal{L}^{(k-1)}(\mathcal{L}^{(k-2)}(\cdots)))\text{,}
\end{equation}
where \mbox{$\mathcal{L}^{(0)} = X$}. In $\mathcal{L}^{(l)}$, the upper index identifies the \mbox{$l$-th} conformal layer, \mbox{$X = \left(x'_1, x'_2, \cdots, x'_{d_{\text{in}}}, x'_o\right)^\intercal$} encodes the input data \mbox{$x \in \mathbb{R}^{d_{\text{in}}}$}, and \mbox{$Y^{(k)} = \left(y^{\prime (k)}_1, y^{\prime (k)}_2, \cdots, y^{\prime (k)}_{d_{\text{out}}}, y^{\prime (k)}_o\right)^\intercal$} encodes the output \mbox{$y \in \mathbb{R}^{d^{(k)}_{\text{out}}}$}, whose actual coordinates can be computed as \mbox{$y_j = y^{\prime (k)}_j / y^{\prime (k)}_o$}, for \mbox{$j = \{1, 2, \cdots, d^{(k)}_{\text{out}}\}$}.

We expand~(\ref{eq:recursive_function}) to model a sequence of $k$ conformal layers applied to $X$ using tensor form:
\begin{equation}
    \label{eq:conformal_layers}
    Y^{(k)} = \left(L_M^{(k)} + L_T^{(k)} X\right) X\text{,}
\end{equation}
where
\begin{equation}
    \label{eq:L_M}
    L_M^{(k)} = F_M^{(k)} U^{(k)} F_M^{(k-1)} U^{(k-1)} \cdots F_M^{(1)} U^{(1)}
\end{equation}
is a \mbox{$(d^{(k)}_{\text{out}} + 1) \times (d_{\text{in}} + 1)$} sparse matrix, and
\begin{multline}
    \label{eq:L_T}
    L_T^{(k)} = \sum_{l=1}^k\left(F_M^{(k)} U^{(k)} F_M^{(k-1)} U^{(k-1)} \cdots  F_M^{(l+1)} U^{(l+1)}\right) \\
    \left(U^{(1)\intercal} U^{(2)\intercal} \cdots U^{(l)\intercal} F_T^{(l)\intercal} U^{(l)} \cdots U^{(2)} U^{(1)}\right)^\intercal
\end{multline}
is a tensor of size \mbox{$(d^{(k)}_{\text{out}} + 1) \times (d_{\text{in}} + 1) \times (d_{\text{in}} + 1)$} filled with zeros, except for the slice at the bottom, which is a sparse \mbox{$(d_{\text{in}} + 1) \times (d_{\text{in}} + 1)$} matrix. The Supplementary Material includes the algebraic manipulation that turns~(\ref{eq:recursive_function}) into~(\ref{eq:conformal_layers}).

Notice that the $L_M^{(k)}$ and $L_T^{(k)}$ components of~(\ref{eq:conformal_layers}) do not depend on the input data $X$, and they encode a complete sequence of $k$ conformal layers. Therefore, after the weights of the CNN be adjusted through the training process, $L_M^{(k)}$ and $L_T^{(k)}$ can be computed once by associativity using~(\ref{eq:L_M}) and~(\ref{eq:L_T}), and applied afterward to any $X$ to perform inference.

\subsection{Implementation of ConformalLayers}
\label{sec:conformal_layers_implementation}

We have implemented ConformalLayers as an \texttt{nn.Module} of PyTorch~1.8\footnote{Source code at https://github.com/Prograf-UFF/ConformalLayers/}. The \texttt{cl.ConformalLayers} module behaves like an \texttt{nn.Sequential} module, and its current version accepts submodules that mimic the behavior of \texttt{nn.Conv1d}, \texttt{nn.Conv2d}, \texttt{nn.Conv3d}, \texttt{nn.AvgPool1d}, \texttt{nn.AvgPool2d}, \texttt{nn.AvgPool3d}, \texttt{nn.Dropout}, and \texttt{nn.Flatten}, in addition to the \texttt{cl.ReSPro} module. The configuration of existing modules includes stride, zero padding, and dilation whenever necessary.

Fig.~\ref{fig:algorithm} shows the algorithm of the \texttt{forward} function implemented by \texttt{cl.ConformalLayers}. During the training process, the \texttt{cl.ConformalLayers} module uses the native PyTorch implementation of supported submodules to compose the network (line~\ref{alg:native_process}). The only exception is the \texttt{cl.ReSPro} module, which is implemented according to~(\ref{eq:respro_without_tensors}), and the computation of the coefficient $y^{\prime (k)}_o$ in all submodules. Once the training process is completed, the \texttt{cl.ConformalLayers} module builds an internal cache containing the sparse matrix $L_M^{(k)}$~(\ref{eq:L_M}) and the sparse matrix representing the slice at the bottom of $L_T^{(k)}$~(\ref{eq:L_T}). This cache has to be updated only if the user decides to retrain the conformal layers (see lines~\ref{alg:invalidade_cache}, \ref{alg:update_cache}, and \ref{alg:validade_cache}). As the value of $y^{\prime (k)}$ can be different from $1$, the result of applying the sequence of submodules to the input data needs to be divided by $y^{\prime (k)}$ to obtain the correct interpretation of the output data (lines~\ref{alg:homogeneous_normalization_1} and~\ref{alg:homogeneous_normalization_2}). Due to the sparse nature of $L_M^{(k)}$ and $L_T^{(k)}$, the expression in line~\ref{alg:computing_Y} is evaluated using only two sparse matrix-vector multiplications, one dot product of vectors, one addition, and one multiplication.

The calculation of sparse tensors $U^{(l)}$ used in the computation of $L_M^{(k)}$ and $L_T^{(k)}$ is implemented using the Minkowski Engine library\footnote{Minkowski Engine: \url{https://github.com/NVIDIA/MinkowskiEngine}} as PyTorch~1.8 does not implement the application of its neural network modules on sparse tensors. The trick used to convert a sequence of linear operations into an $U^{(l)}$ matrix is transforming a sparse identity tensor $I^{(l)}$ by the Minkowski Engine modules that we adapted to behave like PyTorch modules. $I^{(l)}$ is built by stacking a sequence of $d_{\text{in}}^{(l)}$ sparse tensors along the first (batch) dimension. Each of the stacked tensors has the volume expected as input to the $l$-th sequence of linear operations and includes a single $1$ entry identifying which coefficient of the input data is represented by this tensor. The resulting sparse tensor has size \mbox{$d_{\text{in}}^{(l)} \times d_{\text{out}}^{(l)}$}, which correspond to the transposed version of matrix $U^{(l)}$.

The module \texttt{cl.ReSPro} accepts the user to explicitly enter the value of its $\alpha^{(l)}$ argument while defining the network architecture or leave it for the \texttt{cl.ConformalLayers} module estimate the $\alpha^{(l)}$ value. For doing so, we need to keep the tracking of the maximum distance from the origin the point interpretation of the result of the linear operations in the $l$-th conformal layer may have. Before applying the $U^{(l)}$ matrix (which encodes the linear operations), it is reasonable to assume that the $L^2$-norm of the layer's input point is $1$, as long as we set the $x'_o$ coordinate of the input data to the Euclidean distance of $x$ to the origin (line~\ref{alg:input_normalization} in Fig.~\ref{fig:algorithm}). By doing so, all input vector $X$ will encode a point $1$ unit away from the origin, and, by definition, the vector resulting from the application of ReSPro in the previous layer (\emph{i.e.,}~$Y^{(l-1)}$, for \mbox{$l > 1$}) encodes points distant up to $1$ unit from the origin.

\begin{figure}[!t]
    \footnotesize
    \centering
    \rule{\columnwidth}{1pt}\vspace{-2ex}
    \begin{algorithmic}[1]
        \REQUIRE \texttt{input}, a tensor representation of the input data 
        \STATE $x \gets$ \texttt{input} reshaped as a point with $d_{\text{in}}$ coordinates
        \STATE $x'_o \gets \lVert x\rVert_2$ \label{alg:input_normalization}
        \IF{this \texttt{cl.ConformalLayers} module is in training}
            \STATE Set the cache as invalid \label{alg:invalidade_cache}
            \STATE $\text{\texttt{output}}, y^{\prime (k)}_o \gets$ the tensor and the extra coefficient resulting from processing \texttt{input} and $x'_o$ using the sequence of submodules \label{alg:native_process}
            \STATE $\text{\texttt{output}} \gets \text{\texttt{output}} / y^{\prime (k)}_o$ \label{alg:homogeneous_normalization_1}
        \ELSE
            \IF{the cache is invalid}
                \STATE Compute and store $L_M^{(k)}$ and $L_T^{(k)}$ in the cache \label{alg:update_cache}
                \STATE Set the cache as valid \label{alg:validade_cache}
            \ELSE
                \STATE Get $L_M^{(k)}$ and $L_T^{(k)}$ from the cache
            \ENDIF
            \STATE $X \gets \left(x_1, x_2, \cdots, x_{d_{\text{in}}}, x'_o\right)^\intercal$
            \STATE $Y^{(k)} \gets (L_M^{(k)} + L_T^{(k)} X) X$ \label{alg:computing_Y}
            \STATE $y^{(k)} \gets (y^{\prime (k)}_1, y^{\prime (k)}_2, \cdots, y^{\prime (k)}_{d_{\text{out}}}) / y^{\prime (k)}_o$ \label{alg:homogeneous_normalization_2}
            \STATE $\text{\texttt{output}} \gets y^{(k)}$ reshaped as the expected output data
        \ENDIF
        \RETURN \texttt{output}
    \end{algorithmic}
    \rule{\columnwidth}{1pt}
    \caption{The \texttt{cl.ConformalLayers}'s \texttt{forward} function.}
    \label{fig:algorithm}
\end{figure}

But after applying $U^{(l)}$, the maximum distance of the point given to the \texttt{cl.ReSPro} module may change from $1$ to any value, depending on the composition of $U^{(l)}$. Therefore, our implementation needed to deal with each case, progressively updating the maximum distance as the upper limit for the $L^2$-norm that each operation may produce. In the end, $\alpha^{(l)}$ is set to the upper limit of the distance between points emitted by the last operation in $U^{(l)}$ and the origin of the Cartesian space.

For \texttt{nn.AvgPool\textit{N}d}, \texttt{nn.Dropout}, and \texttt{nn.Flatten}, the upper limit for the distance of the resulting point to the origin is equal to the upper limit given as input. For \texttt{nn.Conv\textit{N}d} without bias, the Young's inequality~\cite{young1912multiplication} defines the boundaries of the convolution operator~$\ast$~as:
\begin{equation}
    \label{eq:young_convolution_inequality}
    \lVert g \ast h\rVert_r\ \leq\ \lVert g\rVert_p\ \lVert h\rVert_q\text{, subject to } \frac{1}{p} + \frac{1}{q} = \frac{1}{r} + 1\text{,}
\end{equation}
where $g$ and $h$ denote two discrete signals, and \mbox{$\lVert\cdot\rVert_{t}$} denotes the $L^t$-norm. In our case, we use \mbox{$p = 2$}, since we have the $L^2$-norm of the input vector of the layer, \mbox{$q = 1$}, which corresponds to the $L^1$-norm of the convolution kernel, and \mbox{$r = 2$} for estimating the upper bound for the $L^2$-norm of the output.~The usage of ConformalLayers is transparent to the PyTorch user.


\section{Experiments and Results}
\label{sec:experiments}

\begin{figure*}[!t]
    \centering
    ~\hfill
    \begin{subfigure}[b]{0.19\textwidth}
        \centering
        \includegraphics[width=\textwidth]{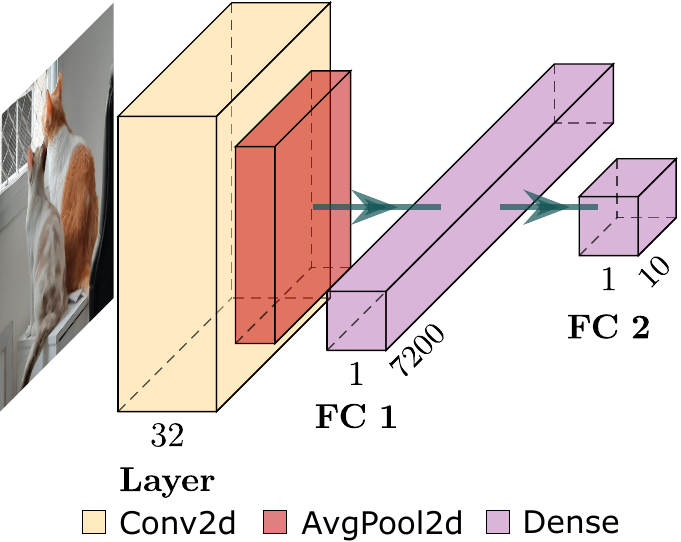}
        \caption{\texttt{BaseLinearNet}}
    \end{subfigure}
    \hfill
    \vrule\
    \hfill
    \begin{subfigure}[b]{0.19\textwidth}
        \centering
        \includegraphics[width=\textwidth]{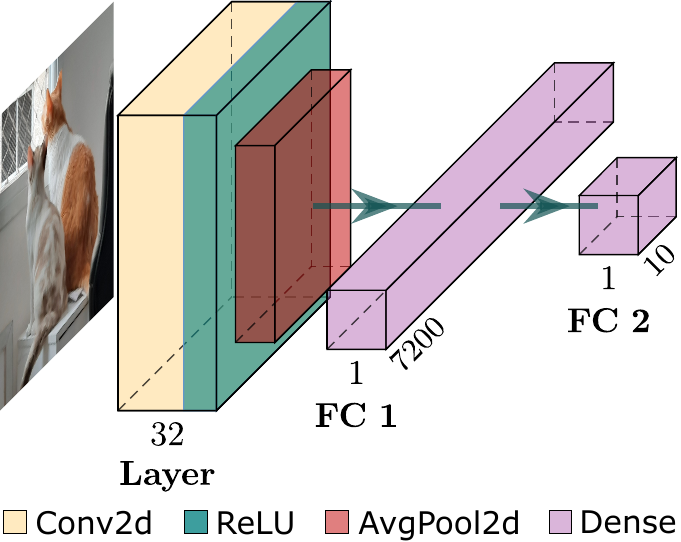}
        \caption{\texttt{BaseReLUNet}}
    \end{subfigure}
    \hfill
    \vrule\
    \hfill
    \begin{subfigure}[b]{0.19\textwidth}
        \centering
        \includegraphics[width=\textwidth]{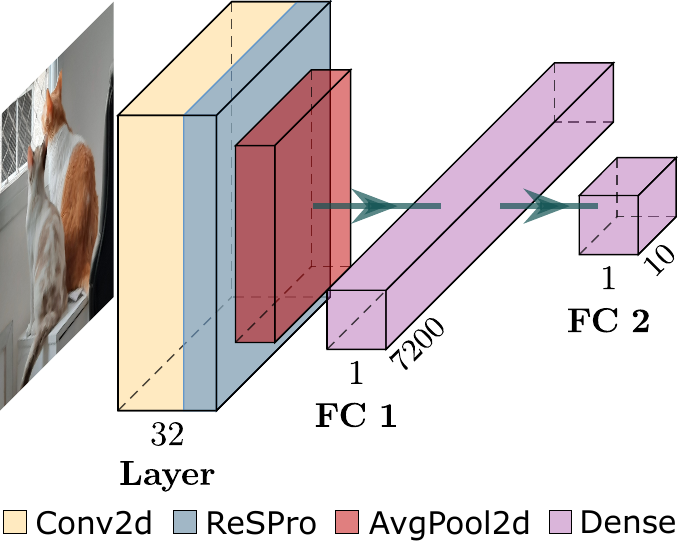}
        \caption{\texttt{BaseReSProNet}}
    \end{subfigure}
    \hfill~
    \caption{Baseline CNNs used in Experiment~I.}
    \label{fig:baseline_architecture}
\end{figure*}

We have performed experiments to assess classification accuracy, memory footprint, and inference time of CNNs implemented using ConformalLayers and their counterparts using typical non-associative layers. For accuracy evaluation~we used MNIST, Fashion-MNIST, and CIFAR-10 datasets. For memory footprint and inference time assessment, we generated large datasets comprised of random RGB images since the classification capacity of the networks are not being compared in these experiments. For Experiments I, II and III~we used an Intel Xeon E5-2698 v4 CPU with $2.2$Ghz with $512$Gb of RAM and 8 GPUs NVIDIA Tesla P100-SXM2 with $16$Gb of memory each. We ran Experiment IV in an Intel Core i7-4770 CPU with $3.4$Ghz with $20$Gb of RAM and a GPU NVIDIA GTX 1050~Ti with $4$Gb of memory. All the experiments were performed inside Docker containers. The number of visible GPUs was set to one even when more GPUs were available.

For each CNN used for image classification, we performed hyperparameter optimization via a Bayesian approach~\cite{bergstra2013making} assuming validation accuracy as metric and Hyperband~\cite{li2017hyperband} as stopping criteria, with arguments $\texttt{max\_iter} = 50$, $s = 2$, and $\eta = 3$. Refer to Supplementary Material for the hyperparameter search space, the graphical analysis produced by the hyperparameter optimization procedure, and the hyperparameter values selected for each CNN/dataset pair. In all experiments, we set the ConformalLayers to automatically estimated the $\alpha$ values used by the ReSPro activation function.

\subsection{Experiment I -- Linear Baseline}
\label{sec:experiment_1}

This experiment aims to measure the performance of three baseline architectures in terms of classification accuracy. The CNNs used in this experiment are very simple to emphasize the effect of the activation function on the result. Fig.~\ref{fig:baseline_architecture} illustrates those architectures. The three CNNs expect \mbox{$32 \times 32$} RGB images as input, include one stage for feature extraction and one fully connected layer with bias to classify the flatted features into one of the $10$ classes defined by the MNIST, Fashion-MNIST, and CIFAR-10 datasets. The feature extraction stage of \texttt{BaseLinearNet} is composed of a convolutional layer with bias follower by average pooling. Notice that this architecture includes only linear layers. \texttt{BaseReLUNet} extends \texttt{BaseLinearNet} by including a ReLU activation function after the convolution. Both CNNs described so far are implemented using native PyTorch modules. In contrast, the feature extraction stage of \texttt{BaseReSProNet} is implemented using ConformalLayers, including one convolutional layer with no bias, the ReSPro activation function, and average pooling. The convolutional kernels in all CNNs have size \mbox{$3 \times 3$}, $32$ output channels, no padding, and stride $1$. The pooling kernels have size \mbox{$2 \times 2$}, no padding and stride~$1$.

The classification results are presented in Table~\ref{tab:baseline_results}. The \texttt{BaseReSProNet} provides higher accuracy when compared to \texttt{BaseLinearNet}, with improvement ranging from $1.36\%$ to $2.17\%$, and lower accuracy when compared to \texttt{BaseReLUNet}, with decrease from $1.95\%$ to $7.61\%$. The comparison to \texttt{BaseLinearNet} reinforces that our activation function makes \texttt{BaseReSProNet} more interesting than a simple linear architecture. The $\alpha$ parameter of ReSPro guarantees the non-linearity needed for model learning. An $\alpha$ value much higher than the maximum $L^2$-norm expected for the input data would cause ReSPro to behave as a linear mapping. According to these experiments, the strategy adopted to automatically estimate $\alpha$ prevents \texttt{BaseReSProNet} from behaving like \texttt{BaseLinearNet}. On the other hand, as the complexity of the dataset increases, we noticed that the accuracy improvement of \texttt{BaseReSProNet} decreases fast, when compared to \texttt{BaseReLUNet}. ReLU seems to provide a better accuracy when compared to ReSPro. This difference suggests a drawback of shallow neural networks using ReSPro to model the complexity of the dataset.

\subsection{Experiment II -- LeNet vs. LeNetCL}
\label{sec:experiment_2}

\begin{table}[!t]
    \caption{Validation accuracy of baseline classification CNNs.}
    \label{tab:baseline_results}
    \centering
    \begin{tabular}{lccc}
        \toprule
        & MNIST & Fashion-MNIST & CIFAR-10 \\
        \midrule
        \texttt{BaseLinearNet} & 92.12\% & 82.19\% & 40.50\% \\
        \texttt{BaseReLUNet} & 97.24\% & 85.90\% & 49.47\% \\
        \texttt{BaseReSProNet} & 94.29\% & 84.01\% & 41.86\% \\
        \bottomrule
    \end{tabular}
\end{table}

\begin{figure}[!t]
    \centering
    \includegraphics[width=0.85\columnwidth]{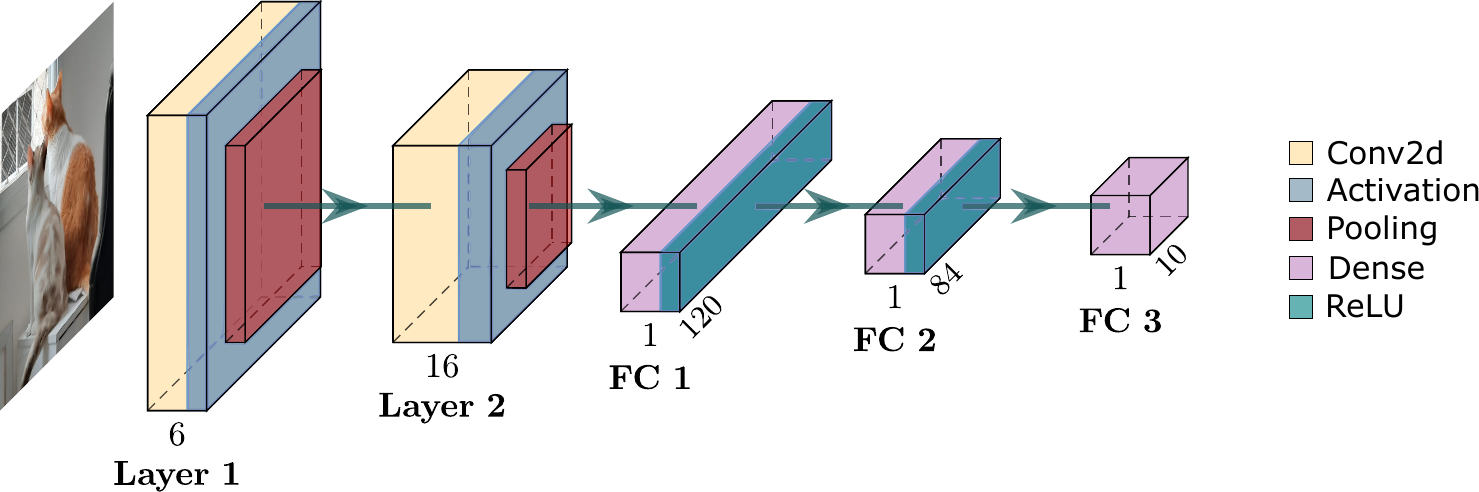}
    \caption{\texttt{LeNetCL}/\texttt{LeNet} architecture used in Experiment~II.}
    \label{fig:lenet_architecture}
\end{figure}

To check if the issue observed on learning the complexity of the dataset is mitigated as we increase the number of layers, we ran an experiment to compared the accuracy of CNNs based on the LeNet-5 architecture~\cite{lecun-01a} (Fig.~\ref{fig:lenet_architecture}). The \texttt{LeNet} CNN consists of layers configured as proposed in the original architecture. The differences are that max-pooling replaced the original pooling function with learned weight and bias, and ReLU replaced the sigmoid activation function. The other CNN, \texttt{LeNetCL}, implements the feature extraction stage (\emph{i.e.,}~layers 1 and 2) using ConformalLayers. Therefore, the use of biases in the convolutions were disabled, and those layers included ReSPro, and average pooling.

Classification results on MNIST, Fashion-MNIST, and CIFAR-10 datasets are reported in Table~\ref{tab:lenet_results}. As it can be noticed, \texttt{LeNetCL} shows a significant improvement when compared to \texttt{BaseReSProNet} in Table~\ref{tab:baseline_results}, mainly because the neural network is deeper and, therefore, able to model more complex data. In addition, the differences in the classification accuracy obtained with the usual implementation of LeNet-5 and \texttt{LeNetCL} are smaller than the differences between \texttt{BaseReLUNet} and \texttt{BaseReSProNet} in all datasets, ranging from $1.01\%$ (Fashion-MNIST) to $5.78\%$ (CIFAR-10).

The relatively small drawback on the accuracy of CNNs built with ReSPro and ReLU is mostly due to the gap between the $\alpha$ value and the upper bound of $L^2$-norm, and may~be acceptable in scenarios where the memory footprint~and computational cost of typical CNNs limit their application.

\subsection{Experiment III -- Network Depth vs. Inference Time}
\label{sec:experiment_3}

In this experiment, we assessed the impact of the depth-independence property of the ConformalLayers on inference time. This property comes from the fact that~(\ref{eq:conformal_layers}) require fixed amounts of memory and arithmetic operations to perform inference, regardless of the number of layers in the network.

Batches including $64$ random RGB images having \mbox{$32 \times 32$} pixels with intensities sorted from a uniform distribution were used as input in this experiment. We also set the weights of the compared CNNs to uniformly distributed random values since the objective is to measure the computation time of the solutions instead of analyzing their classification capabilities.

\begin{figure}[!b]
    \centering
    \includegraphics[width=0.9\columnwidth]{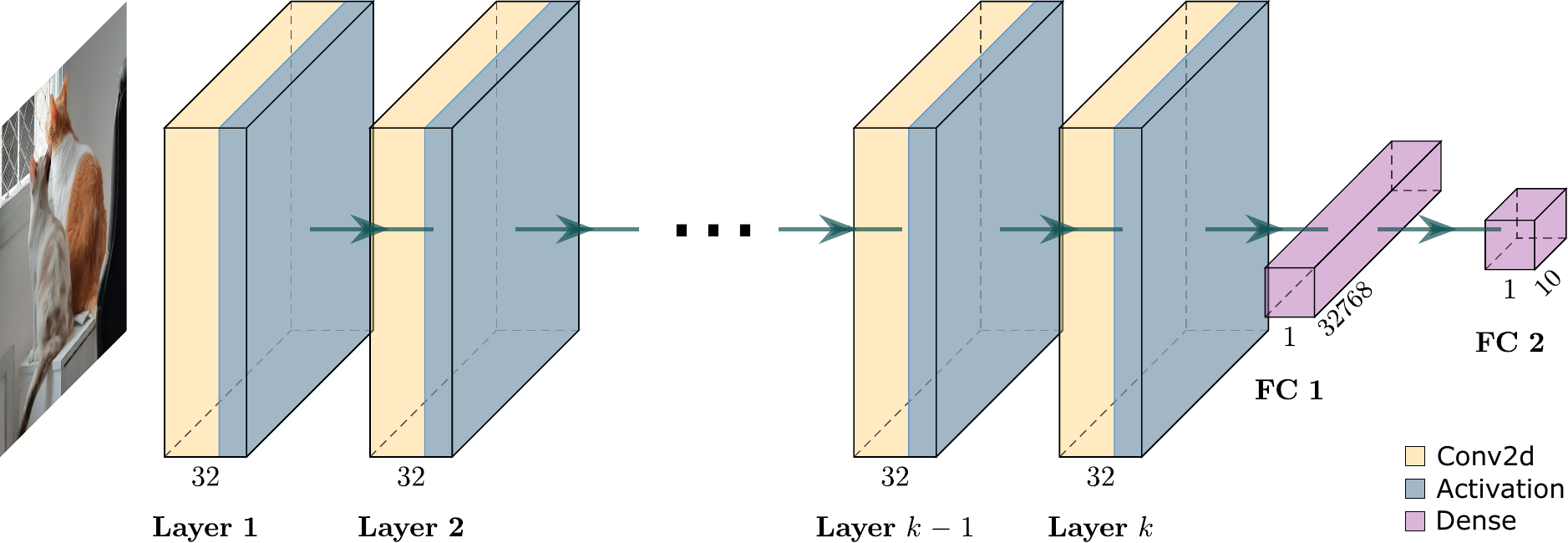}
    \caption{\texttt{DkNetCL}/\texttt{DkNet} architecture used in Experiment~III.}
    \label{fig:dknet_architecture}
\end{figure}

Fig.~\ref{fig:dknet_architecture} illustrates the architecture of the CNNs used in this experiment. They comprise sequences of convolutional layers with kernels of size \mbox{$3 \times 3$}, $32$ output channels, stride of $1$, and padding of $1$ to keep input and output with the same size. We placed an activation function following each convolution layer. After the first $k$ layers, both CNNs include two fully connected layers, with bias, that produce vectors with, respectively, $32.768$ and $10$ entries. The differences between \texttt{DkNetCL} and \texttt{DkNet} are that the former uses ReSPro as activation function, no bias in the convolutions, and implements the first $k$ layers as ConformalLayers. The latter uses native PyTorch modules, ReLU, and includes bias in the convolutions.

\begin{figure}[!b]
    \centering
    \includegraphics[width=0.6\columnwidth]{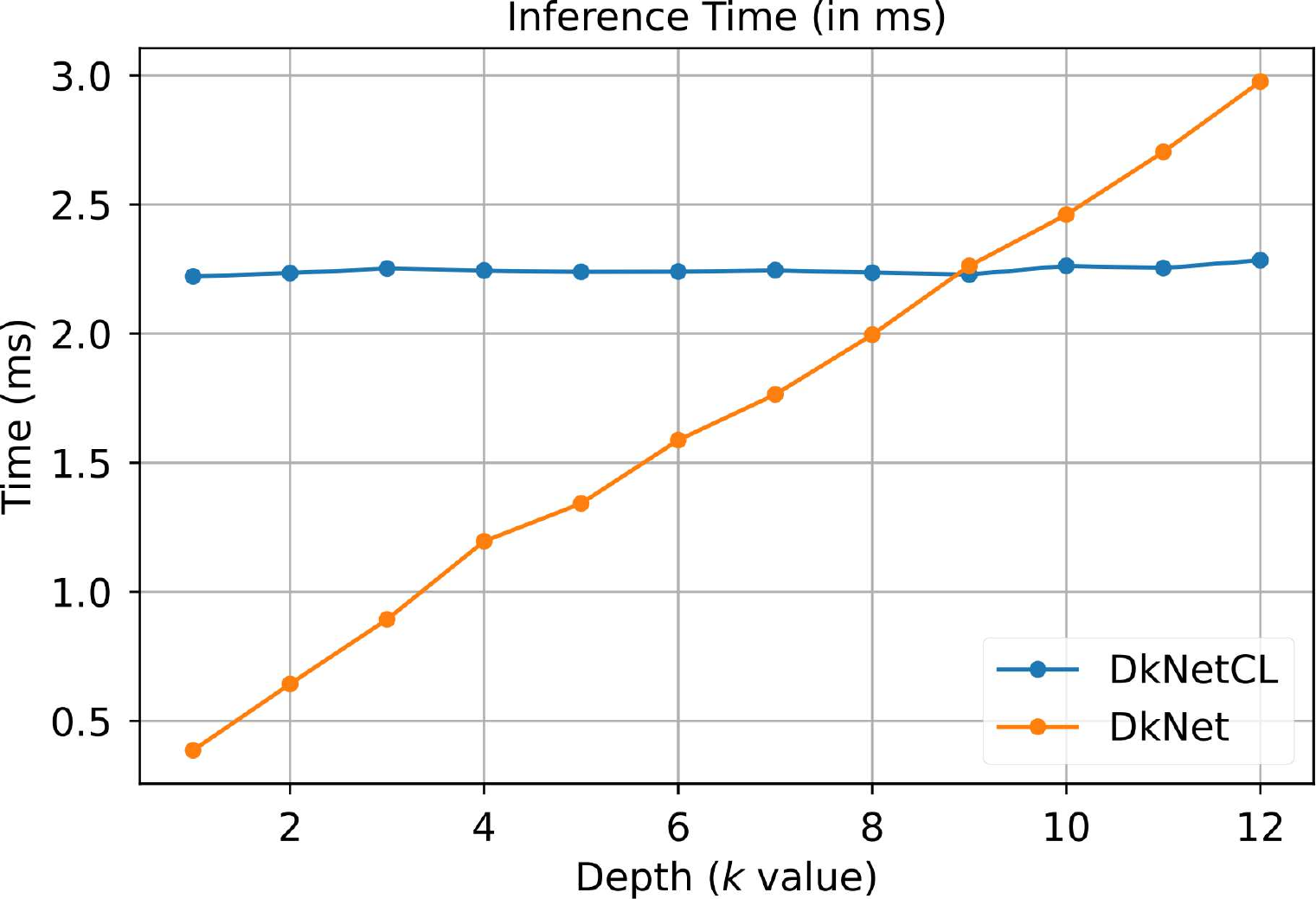}
    \caption{Inference times for \texttt{DkNetCL} and \texttt{DkNet}.}
    \label{fig:depth_analysis}
\end{figure}

The average inference time of $100$ executions of \texttt{DkNetCL} and \texttt{DkNet} for different depths $k$ is presented in Fig.~\ref{fig:depth_analysis}. Our first observation relies on the constant inference times of \texttt{DkNetCL} as we increase the depth of the neural network. \texttt{DkNet}, on the other hand, shows inference time as a linear monotonically crescent curve. For \mbox{$k = 1$}, \texttt{DkNet} is almost $6\times$ faster than \texttt{DkNetCL}. This apparent drawback, however, is quickly overcome as the neural networks become deeper. For \mbox{$k \geq 9$}, the ConformalLayers-based CNN becomes more profitable than the one using non-associative layers.

We believe that much of the processing time required in the inference performed by networks based on ConformalLayers is due to the computational inefficiency of existing sparse matrix libraries compared to the same procedures implemented to dense matrices. Unfortunately, the size of the matrices involved prevents the use of dense matrices in our solution. But it is noted that recent versions of libraries like PyTorch have paid particular attention to operations with sparse arrays, gradually improving the performance of the available implementations.

\subsection{Experiment IV -- Batch Size vs. Inference Time}
\label{sec:experiment_4}

\begin{figure}[!t]
    \centering
    \includegraphics[width=0.85\columnwidth]{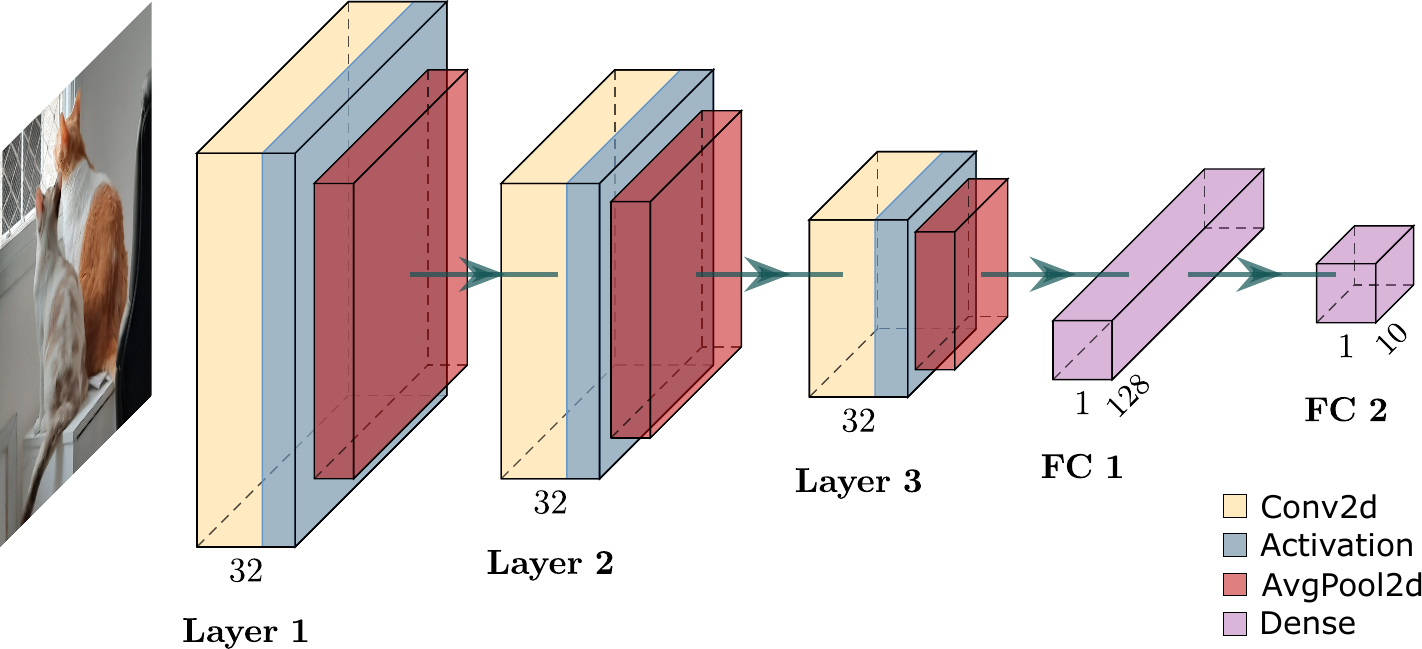}
    \caption{\texttt{D3ModNetCL}/\texttt{D3ModNet} CNNs from Experiment~IV.}
    \label{fig:d3modnet_architecture}
\end{figure}

This experiment compares the processing time and the supported number of images processed simultaneously using a ConformalLayers-based solution and its conventional counterpart while performing inference. For this experiment, we defined the \texttt{D3ModNetCL} and \texttt{D3ModNet} networks, which implement the architecture presented in Fig.~\ref{fig:d3modnet_architecture}. As in previous experiments, the networks expect \mbox{$32 \times 32$} RGB images as input and the difference between \texttt{D3ModNetCL} and \texttt{D3ModNet} is in the choice of the activation function, the use of bias in convolutions, and the possibility of applying associativity in the first three layers. Both CNNs have fixed depth, perform \mbox{$2 \times 2$} average pooling after the activation function, pooling and convolutions have no padding, and FC layers have bias.

The average processing times resulting from $100$ executions of each compared CNN are presented in Fig.~\ref{fig:inference_D3NetCL_D3Net}, where one can notice that \texttt{D3ModNetCL} performs inference faster than \texttt{D3ModNet}. The zoomed-in portion of the curves shows an interesting behavior: the inference time of \texttt{D3ModNetCL} increases in steps, where each step has a length of 32 batches. This is related to the size of the thread block of the GPU used in this experiment. In practice, the size of thread blocks defines the number of cycles needed to perform some calculation.

\begin{table}[!t]
    \caption{Validation accuracy of \texttt{LeNet} and \texttt{LeNetCL}.}
    \label{tab:lenet_results}
    \centering
    \begin{tabular}{lccc}
        \toprule
        & MNIST & Fashion-MNIST & CIFAR-10 \\
        \midrule
        \texttt{LeNet} & 98.87\% & 90.13\% & 59.05\% \\
        \texttt{LeNetCL} & 97.33\% & 89.12\% & 53.27\% \\
        \bottomrule
    \end{tabular}
\end{table}

\begin{figure}[!t]
    \centering
    \includegraphics[width=0.6\columnwidth]{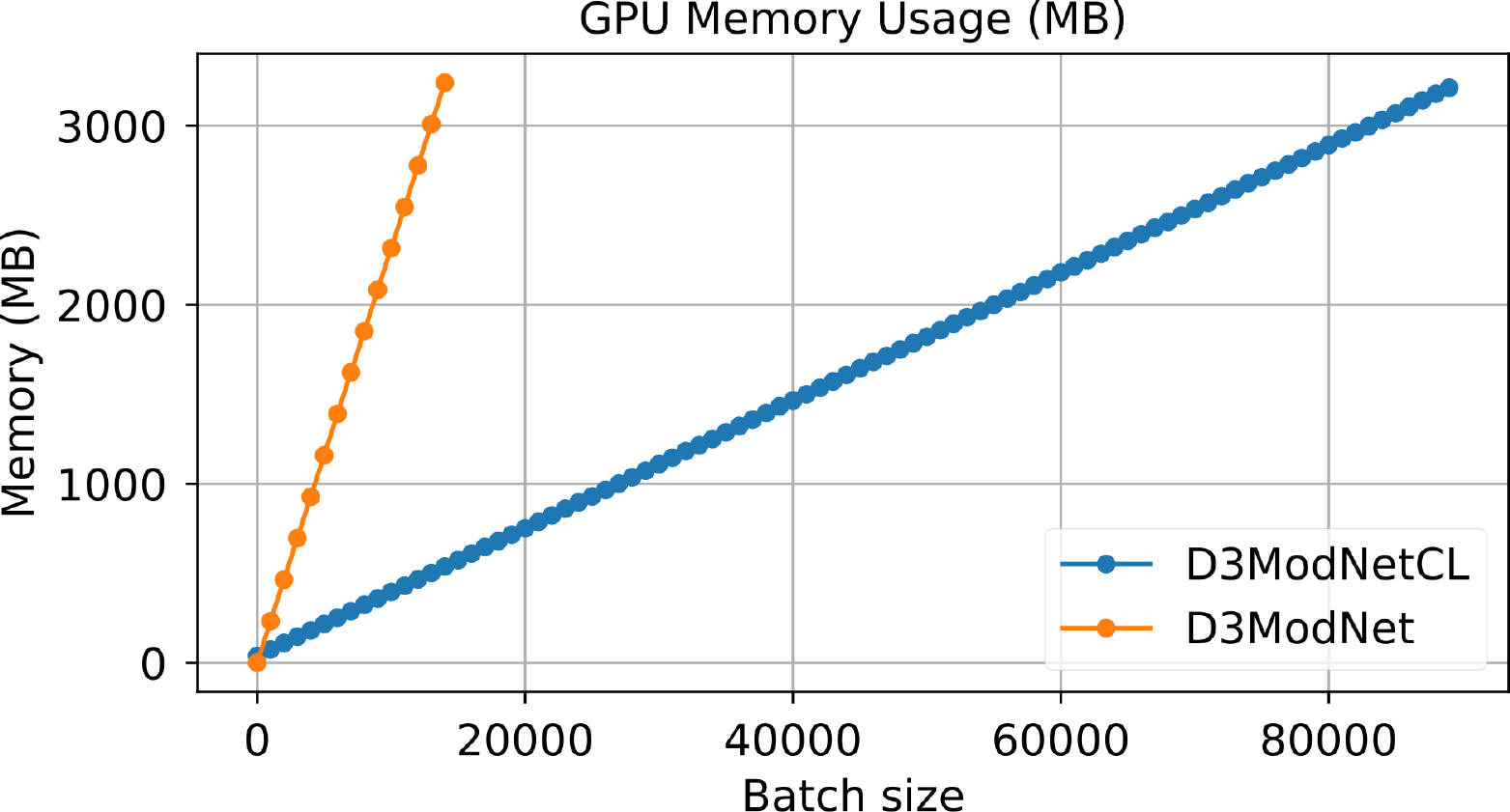}
    \caption{Memory footprint for \texttt{D3ModNetCL} and~\texttt{D3ModNet}.}
    \label{fig:gpu_memory_usage}
\end{figure}

Another interesting observation can be done if we analyze Fig.~\ref{fig:inference_D3NetCL_D3Net} alongside Fig.~\ref{fig:gpu_memory_usage}. The GPU used has $4$GB of memory. As one may notice in Fig.~\ref{fig:gpu_memory_usage}, \texttt{D3ModNet}'s memory usage line has a higher slope when compared to \texttt{D3ModNetCL}. As a result, the approach based on non-associative layers consumes the full GPU memory quickly. In contrast, the ConformalLayers-based approach supports larger batches before using all the memory. It is because the feature map resulting from each operation in \texttt{D3ModNet} has to be stored in memory to be passed as input to the next operation. Hence, \texttt{D3ModNet}'s curve in Fig.~\ref{fig:inference_D3NetCL_D3Net} shows that the maximum batch size for this architecture in this GPU is $14$K, while the \texttt{D3ModNetCL} supports up to $89$K images simultaneously.

We used linear regression in Fig.~\ref{fig:inference_D3NetCL_D3Net} to extrapolate the memory bottleneck of \texttt{D3ModNet} and estimate its inference time with more than $14$K images. The extrapolation allows the comparison of both approaches under the maximum capacity of \texttt{D3ModNetCL}. Fig.~\ref{fig:inference_D3NetCL_D3Net} shows that our approach is about \mbox{$1.16\times$} times faster than the non-associative CNN. Such improvement and memory saving suggest that the presented technique is feasible for devices with limited capabilities.


\section{Conclusions and Future Work}
\label{sec:conclusion}

We presented the ReSPro, a novel non-linear and differentiable activation function that can be fully encoded as matrices and \mbox{rank-$3$} tensors, whose product satisfies the associative law. In our experiments, we did not notice ReSPro suffering from vanishing or exploding gradient, like other S-shaped functions, probably because it behaves like element-wise activation followed by normalization instead of performing simple element-wise transformations. We also presented a new neural network back end called ConformalLayers. To the best of our knowledge, ConformalLayers is the first non-linear sequential CNN with associative layers. The combination of layers by associativity has several advantages in terms of computational efficiency during inference.

The current implementation of ConformalLayers does not support channel groups nor bias in convolutions, and does not implement transposed convolution and fully connected layers. We are working on adding these features to our framework.

Although this paper presents ConformalLayers as an architecture for sequential CNNs, we believe that we can expand the concept to non-sequential networks. One possibility is, before inference, to traverse the graph defined by the layers of non-sequential CNNs in a depth-first fashion and apply the associativity of the ConformalLayers to the paths found.

We hope that our original ideas lead to new lines of investigation. Possible directions of future exploration include~the proposition of other activation functions, and the study of the data encoded by the matrix $L_M^{(k)}$ and \mbox{rank-$3$} tensor $L_T^{(k)}$.


\section*{Acknowledgments}

This work was partially supported by CNPq (\mbox{311.037/2017-8}) and FAPERJ (\mbox{E-26/202.718/2018}) agencies.

\bibliographystyle{IEEEtran}
\bibliography{references}

\end{document}


\title{ConformalLayers: A non-linear sequential neural network with associative layers\\\Large{-- Supplementary Material --}}

\author{\IEEEauthorblockN{Eduardo Vera Sousa, Leandro A. F. Fernandes, Cristina Nader Vasconcelos}
    \IEEEauthorblockA{Instituto de Computação, Universidade Federal Fluminense (UFF)\\
    Niterói, Rio de Janeiro, Brazil -- ZIP 24210--346\\
    Email: \{eduardovera, laffernandes, crisnv\}@ic.uff.br}
}

\maketitle

\begin{tikzpicture}[remember picture, overlay]%
\node at ($(current page.north) + (0,-0.5in)$) {Best Paper on Pattern Recognition and Related Field at SIBGRAPI 2021 -- 34th Conference on Graphics, Patterns and Images};%
\end{tikzpicture}%

The Supplementary Material is structured as follows. In Section~\ref{sec:respro_tensors}, we present the algebraic manipulation that takes the ReSPro function from its formulation using the geometric algebra of the conformal model for Euclidean geometry~\cite{dorst2010geometric} to tensor algebra. Section~\ref{sec:typical_linear_layers_and_operations} shows the matrix representation of typical linear layers and other operations in CNNs. The algebraic manipulation that expands a sequence of calls to $\mathcal{L}^{(l)}$ functions, for \mbox{$l \in \{1, 2, \cdots, k\}$}, to produce the tensor representation of the ConformalLayers is presented in Section~\ref{sec:confomal_layers_tensors}. The hyperparameter values selected for each CNN and dataset used in Experiments~I and~II is presented in Section~\ref{sec:hyperparamter_values}.  Sections~\ref{sec:sweep_baseline} and~\ref{sec:sweep_lenet} present the hyperparameter sweep. Please refer to the paper for details about the experiments and discussion on the results.

\section{ReSPro: From Geometric Algebra to Tensors}
\label{sec:respro_tensors}

We represent the discrete input data as a point \mbox{$x = \left(x_1, x_2, \cdots, x_d\right) \in \mathrm{R}^{d}$}. We include an extra dimension in $\mathrm{R}^{d}$ and embed $\mathrm{R}^{d+1}$ in a \mbox{$(d + 3)$}-dimensional space with basis vectors \mbox{$\{e_1, e_2, \cdots, e_{d+1}, e_o, e_\infty\}$} and metric matrix:
\begin{equation*}
	\begin{array}{c|c c c c c c}
		\cdot & e_1 & e_2 & \cdots & e_{d+1} & e_o & e_\infty \\
		\hline
		e_1 & 1 & 0 & \cdots & 0 & \phantom{-}0 & \phantom{-}0 \\
		e_2 & 0 & 1 & \cdots & 0 & \phantom{-}0 & \phantom{-}0 \\
		\vdots & \vdots & \vdots & \ddots & \vdots & \phantom{-}\vdots & \phantom{-}\vdots \\
		e_{d+1} & 0 & 0 & \cdots & 1 & \phantom{-}0 & \phantom{-}0 \\
		e_o & 0 & 0 & \cdots & 0 & \phantom{-}0 & -1 \\
		e_\infty & 0 & 0 & \cdots & 0 & -1 & \phantom{-}0
	\end{array}
\end{equation*}
where $\cdot$ denotes the vector inner product. In other words, we assume the conformal model for Euclidean geometry to work with our data. In this model, the finite point $x$ is represented by the vector:
\begin{equation}
    V = x'_1 e_1 + x'_2 e_2 + \cdots + x'_d e_d + x'_o e_o + \frac{x'_o}{2} \sum_{i=1}^{d} (x_i)^{2} e_\infty \text{,}
\end{equation}
where \mbox{$x_i = x'_i / x'_o$}, for \mbox{$i \in \{1, 2, \cdots, d + 1\}$}, and \mbox{$x'_o \neq 0$}. By definition, we set \mbox{$x_{d+1} = 0$}. The extra dimensions $e_o$ and $e_\infty$ are geometrically interpretable as the point at the origin and the point at infinity, respectively.

We construct a geometric algebra over the conformal space. This allows operations on the base space, including reflections, rotations, and translations to be represented using versors of the geometric algebra~\cite{dorst2010geometric}. Therefore, from now on, the mathematical expressions will be written using geometric algebra. The terms in the expressions that follow represent blades or versors encoded as multivectors in the multivector space $\bigwedge\mathbb{R}^{d+3}$. The half-space denotes the geometric product, while $\op$ and $\lcont$ denote, respectively, the outer (wedge) product and the left contraction.

Let $V'$ be a $2$-blade encoding a planar point compute from $V$ and $e_\infty$ using the outer product:
\begin{equation}
    \begin{split}
        V' = V \op e_\infty
        &= \left(x'_1 e_1 + x'_2 e_2 + \cdots + x'_d e_d + x'_o e_o + x'_\infty e_\infty\right) \op e_\infty \\
        &= \left(x'_1 e_1 + x'_2 e_2 + \cdots + x'_d e_d + x'_o e_o\right) \op e_\infty \\
        &= v \op e_\infty + x'_o \left(e_o \op e_\infty\right) \\
        &= \left(v + x'_o e_o\right) \op e_\infty \text{,}
    \end{split}
\end{equation}
where \mbox{$v = x'_1 e_1 + x'_2 e_2 + \cdots + x'_d e_d$}, and \mbox{$x'_\infty = \frac{x'_o}{2} \sum_{i=1}^{d} (x_i)^{2}$}.

The reflection of $V'$ in a hypersphere with center \mbox{$c = \left(0, 0, \cdots, \alpha\right) \in \mathrm{R}^{d+1}$} and radius $\alpha$, for \mbox{$\alpha \geq 0$}, followed by isotropic scaling by a factor of \mbox{$2 / \alpha$}, is done by applying the $3$-versor $T$ to $V'$, which leads to the $2$-blade $V''$ encoding a pair of points where the second point of the pair corresponds to the center of the hypersphere properly transformed by scaling. The versor $T$ is given by:
\begin{equation}
    T = \left(\cosh\left(\frac{1}{2} \log\frac{2}{\alpha}\right) + \sinh\left(\frac{1}{2} \log\frac{2}{\alpha}\right) \left(e_o \op e_\infty\right)\right) \gp \left(\alpha e_{d+1} + e_o\right) \text{,}
\end{equation}
whose inverse is:
\begin{equation}
    T^{-1} = \left(\frac{1}{\alpha} e_{d+1} + \frac{1}{\alpha^2} e_o\right) \gp \left(\cosh\left(\frac{1}{2} \log\frac{2}{\alpha}\right) - \sinh\left(\frac{1}{2}\log\frac{2}{\alpha}\right) \left(e_o \op e_\infty\right)\right)\text{.}
\end{equation}
The resulting pair of points $V''$ is computed using the versor product of $V'$ by $T$:
\begin{equation}
    \begin{split}
        V'' = T \gp V' \gp T^{-1}
        &= T \gp \left(V \op e_\infty\right) \gp T^{-1} \\
        &= \left(-T \gp V \gp T^{-1}\right) \op \left(-T \gp e_\infty \gp T^{-1}\right) \\
        &= \frac{2}{\alpha} \left(v \op e_{d+1}\right) + \frac{1}{\alpha} \left(v \op e_o\right) + \frac{2}{\alpha} \left(v \op e_\infty\right) - x'_o \left(e_{d+1} \op e_o\right) + x'_o \left(e_o \op e_\infty\right) \text{.}
    \end{split}
\end{equation}

By contracting $e_\infty$ on $V''$ one gets the dual of the perpendicular bisector of the pair of point $V''$:
\begin{equation}
    \begin{split}
        H = e_\infty \lcont V''
        &= e_\infty \lcont \left(\frac{2}{\alpha} \left(v \op e_{d+1}\right) + \frac{1}{\alpha} \left(v \op e_o\right) + \frac{2}{\alpha} \left(v \op e_\infty\right) - x'_o \left(e_{d+1} \op e_o\right) + x'_o \left(e_o \op e_\infty\right)\right) \\
        &= \frac{1}{\alpha} v - x'_o e_{d+1} - x'_o e_\infty \text{.}
    \end{split}
\end{equation}

The bisector can be used as a ``mirror'' to reflect the known point of the pair to find the unknown point. The computed point is finite and represents the input point transformed by spherical projection followed by scaling. We use the versor product to compute where the point at infinity (\emph{i.e.,}~the known point in the original pair $V'$) was mapped by $T$. Next we use an up-to-scaling versor product to find $V'''$ as the reflection of the transformed point at infinity. The outer product of the resulting point with $e_\infty$ simplifies the expression by computing the flat point at the same location. Finally, the point is projected orthographically to \mbox{$e_1 \op e_2 \op \cdots \op e_d \op e_o$}:
\begin{equation}
    \begin{split}
        V''' &= \textsc{project}\left(\left(-H \gp \left(-T \gp e_\infty \gp T^{-1}\right) \gp H\right) \op e_\infty\right) \\
        &= \textsc{project}\left(\left(-H \gp \left(\frac{2}{\alpha} e_{d+1} + \frac{1}{\alpha} e_o + \frac{2}{\alpha} e_\infty\right) \gp H\right) \op e_\infty\right) \\
        &= \textsc{project}\left(\left(-\frac{2 x'_o}{\alpha^2} v - \frac{2 \left(v \cdot v\right)}{\alpha^3} e_{d+1} - \frac{1}{\alpha} \left(\frac{\left(v \cdot v\right) x'_o}{\alpha^2} + x^{\prime 2}_o\right) e_o - \frac{2 \left(v \cdot v\right)}{\alpha^3} e_\infty\right) \op e_\infty\right) \\
        &= \textsc{project}\left(-\frac{2 x'_o}{\alpha^2} \left(v \op e_\infty\right) - \frac{2 \left(v \cdot v\right)}{\alpha^3} \left(e_{d+1} \op e_\infty\right) - \left(\frac{\left(v \cdot v\right) x'_o}{\alpha^3} + \frac{x^{\prime 2}_o}{\alpha}\right) \left(e_o \op e_\infty\right)\right) \\
        &= -\frac{2 x'_o}{\alpha^2} \left(v \op e_\infty\right) - \left(\frac{\left(v \cdot v\right) x'_o}{\alpha^3} + \frac{x^{\prime 2}_o}{\alpha}\right) \left(e_o \op e_\infty\right) \text{.}
    \end{split}
\end{equation}
The resulting flat point $Y$ is computed by multiplying $V'''$ by $-\alpha^2 / \left(2 x'_o\right)$ to simplify the expression:
\begin{equation}
    \label{eq:Y_using_ga}
    \begin{split}
        Y &= -\frac{\alpha^{2}}{2 x'_o} V''' \\
        &= -\frac{\alpha^{2}}{2 x'_o} \left(-\frac{2 x'_o}{\alpha^2} \left(v \op e_\infty\right) - \left(\frac{\left(v \cdot v\right) x'_o}{\alpha^3} + \frac{x^{\prime 2}_o}{\alpha}\right) \left(e_o \op e_\infty\right)\right) \\
        &= \left(v \op e_\infty\right) + \left(\frac{\left(v \cdot v\right)}{2 \alpha} + \frac{\alpha x'_o}{2}\right) \left(e_o \op e_\infty\right) \\
        &= \left(v + \left(\frac{\left(v \cdot v\right)}{2 \alpha} + \frac{\alpha x'_o}{2}\right) e_o\right) \op e_\infty \text{.}
    \end{split}
\end{equation}
Recall that the interpretation of $V'''$ as a flat point at location $y$ is unchanged if its coefficients are multiplied by a common factor different than zero.

Finally, the flat point $Y$, in~(\ref{eq:Y_using_ga}), can be written using tensors to represent the point $y$ resulting from applying ReSPro to $x$:
\begin{align}
    Y = \begin{pmatrix}
            y'_1 \\
            y'_2 \\
            \vdots \\
            y'_d \\
            y'_o
        \end{pmatrix} = \begin{pmatrix}
            x'_1 \\
            x'_2 \\
            \vdots \\
            x'_d \\
            \frac{\alpha}{2} x'_{o} + \frac{1}{2 \alpha} \sum_{i=1}^{d} (x'_{i})^{2} \\
        \end{pmatrix}
    &= \left(
        \begin{pmatrix}
            1 & 0 & \cdots & 0 & 0 \\ 
            0 & 1 & \cdots & 0 & 0 \\ 
            \vdots & \vdots & \ddots & \vdots & \vdots \\ 
            0 & 0 & \cdots & 1 & 0 \\ 
            0 & 0 & \cdots & 0 & \frac{\alpha}{2} \\ 
        \end{pmatrix} + \begin{pmatrix}
            0 & 0 & \cdots & 0 & 0 \\ 
            0 & 0 & \cdots & 0 & 0 \\ 
            \vdots & \vdots & \ddots & \vdots & \vdots \\ 
            0 & 0 & \cdots & 0 & 0 \\ 
            \frac{x'_1}{2 \alpha} & \frac{x'_2}{2 \alpha} & \cdots & \frac{x'_d}{2 \alpha} & 0 \\ 
        \end{pmatrix}
    \right) 
    \begin{pmatrix}
        x'_1 \\
        x'_2 \\
        \vdots \\
        x'_d \\
        x'_o
    \end{pmatrix} \\
    &= \left(F_M + F_T X\right) X\text{.}
\end{align}

\section{Linear CNN Layers and Other Operations as Matrices}
\label{sec:typical_linear_layers_and_operations}

Each paragraph that follows models a typical linear operation or configuration in CNNs in matrix form. Without loss of generality, we assume that the discrete signals transformed by these matrices are one-dimensional and have a single channel. The presentation of the multidimensional and multi-channel version of these expressions is out of scope of this Supplementary Material.

\paragraph{Data} The data is modeled as vectors in $\mathbb{R}^{d_{x} + 1}$, where $d_{x}$ is the size of the data and ${}^\intercal$~denotes matrix transposition:
\begin{equation}
    X = \left(x'_1, x'_2, \cdots, x'_{d_{x}}, x'_o\right)^\intercal \text{.}
\end{equation}

\paragraph{Weights} The weights are modeled as vectors in $\mathbb{R}^{d_{w} + 1}$, where $d_{w}$ is the size of the kernel:
\begin{equation}
    W = \left(w_1, w_2, \cdots, w_{d_{w}}, 1\right)^\intercal \text{.}
\end{equation}

\paragraph{Padding of Zeros} The matrix \mbox{$P = \left(p_{ij}\right)$} is a constant \mbox{$(d_{x} + 2 \delta_P + 1) \times (d_{x} + 1)$} matrix that models zero padding of $\delta_P$ units, where $\delta_P$ is the number of zeros inserted at each end of the signal represented as a vector in $\mathbb{R}^{d_{x}+1}$:
\begin{equation}
    p_{ij} = \begin{cases}
        1 & \text{, for } ((i - \delta_P) = j \text{ and } i < P_{\text{rows}} \text{ and } j < P_{\text{cols}}) \text{ or } (i = P_{\text{rows}} \text{ and } j = P_{\text{cols}}) \text{,} \\
        0 & \text{, otherwise.} \\
    \end{cases}
\end{equation}

\paragraph{Dilation} The matrix \mbox{$D = \left(d_{ij}\right)$} is a constant \mbox{$(d_{w} + 1) \times ((d_{w} - 1) \delta_D + 2)$} matrix that models dilation with dilation rate $\delta_D$:
\begin{equation}
    d_{ij} = \begin{cases}
        1 & \text{, for } ((i - 1) \delta_D + 1 = j \text{ and } i < D_{\text{rows}} \text{ and } j < D_{\text{cols}}) \text{ or } (i = D_{\text{rows}} \text{ and } j = D_{\text{cols}}) \text{,} \\
        0 & \text{, otherwise.}
    \end{cases}
\end{equation}

\paragraph{Stride} The matrix \mbox{$S = \left(s_{ij}\right)$} is a constant \mbox{$(\left\lfloor\frac{(d_{x} + 2 \delta_{P} - \delta_{D} (d_{w} - 1) - 1)}{\delta_{S}}\right\rfloor + 2) \times (d_{x} - (d_{w} - 1) \delta_{D} + 2 \delta_{P} + 1)$} matrix that model stride with displacement $\delta_{S}$:
\begin{equation}
    s_{ij} = \begin{cases}
        1 & \text{, for } ((i - 1) \delta_{S} = j - 1 \text{ and } i < S_{\text{rows}} \text{ and } j < S_{\text{cols}}) \text{ or } (i = S_{\text{rows}} \text{ and } j = S_{\text{cols}}) \text{,} \\
        0 & \text{, otherwise.}
    \end{cases}
\end{equation}

\paragraph{Convolution} The matrix $M$ that models convolution is computed as a composition of weights, dilation, valid cross-correlation, stride, and padding:
\begin{equation}
    M = S \, \left(W \, D \, C\right) \, P = W \, \left(S \, \left(D \, C\right)^\intercal \, P\right)^\intercal \text{,}
\end{equation}
where the constant rank-$3$ tensor \mbox{$C = \left(c_{ijk}\right)$}, of size $((d_{w} - 1) \delta_{D} + 2) \times (d_{x} - (d_{w} - 1) \delta_{D} + 2 \delta_{P} + 1) \times (d_{x} + 2 \delta_{P} + 1)$, models the valid cross-correlation:
\begin{equation}
    d_{ijk} = \begin{cases}
        1 & \text{, for } (i = k - j + 1 \text{ and } i < C_{\text{dim1}} \text{ and } j < C_{\text{dim2}} \text{ and } k < C_{\text{dim3}}) \text{ or } (i = C_{\text{dim1}} \text{ and } j = C_{\text{dim2}} \text{ and } k = C_{\text{dim3}}) \text{,} \\
        0 & \text{, otherwise.}
    \end{cases}
\end{equation}

\paragraph{Average Pooling} The constant matrix $A$ modeling average pooling is computed as a composition of constant weights, valid cross-correlation, stride, and padding:
\begin{equation}
    A = S \, \left(W \, C\right) \, P = W \, \left(S \, C^\intercal \, P\right)^\intercal \text{,}
\end{equation}
where \mbox{$W = \left(\frac{1}{d_{w}}, \frac{1}{d_{w}}, \cdots, \frac{1}{d_{w}}, 1\right)^\intercal \in \mathrm{R}^{d_{w}+1}$} is a constant vector of weights and $d_{w}$ is the size of the kernel.

\paragraph{Dropout} The \mbox{$(d_{x} + 1) \times (d_{x} + 1)$} diagonal matrix $R = \left(r_{i,j}\right)$ encodes the dropout operation. Its entries are:
\begin{equation}
    r_{ij} = \begin{cases}
        1 & \text{, for } i = j \text{ and } \textsc{rand}() > \delta_{R} \text{,} \\
        0 & \text{, otherwise.}
    \end{cases}
\end{equation}
Here, $\textsc{rand}()$ is a function that generates random values uniformly distributed on the interval \mbox{[0, 1]}, and $\delta_{R}$ is the probability of an element to be zeroed.

\section{ConformalLayers: From Successive Evaluation of $\mathcal{L}^{(l)}$ Functions to Tensors}
\label{sec:confomal_layers_tensors}

For the sake o clarity, we first show the algebraic manipulation that turns:
\begin{equation}
    \label{eq:recursive_function}
    Y^{(k)} = \mathcal{L}^{(k)}(\mathcal{L}^{(k-1)}(\mathcal{L}^{(k-2)}(\cdots)))
\end{equation}
into
\begin{equation}
    Y^{(k)} = \left(L_M^{(k)} + L_T^{(k)} X\right) X
\end{equation}
for \mbox{$k = 3$}, and then define the case for any $k$ by induction.

Let \mbox{$k = 3$}. Equation~(\ref{eq:recursive_function}) expands to:
\begin{equation}
    \begin{split}
        Y^{(3)}
        &= \mathcal{L}^{(3)}\left(\mathcal{L}^{(2)}\left(\mathcal{L}^{(1)}(X)\right)\right) \\
        &= \!\begin{multlined}[t]
            F_{M}^{(3)} U^{(3)} \left(F_{M}^{(2)} U^{(2)} F_{M}^{(1)} U^{(1)} X + \left(F_{M}^{(2)} U^{(2)} \left(U^{(1)\intercal} F_{T}^{(1)\intercal} U^{(1)}\right)^\intercal X\right) X + \left(\left(U^{(1)\intercal} U^{(2)\intercal} F_{T}^{(2)\intercal} U^{(2)} U^{(1)}\right)^\intercal X\right) X\right) \\
            + \left(\left(U^{(3)\intercal} F_{T}^{(3)\intercal} U^{(3)}\right)^\intercal \left(F_{M}^{(2)} U^{(2)} F_{M}^{(1)} U^{(1)} X + \left(F_{M}^{(2)} U^{(2)} \left(U^{(1)\intercal} F_{T}^{(1)\intercal} U^{(1)}\right)^\intercal X\right) X\right.\right. \\
            \left.\left.+ \left(\left(U^{(1)\intercal} U^{(2)\intercal} F_{T}^{(2)\intercal} U^{(2)} U^{(1)}\right)^\intercal X\right) X\right)\right) \\
            \left(F_{M}^{(2)} U^{(2)} F_{M}^{(1)} U^{(1)} X + \left(F_{M}^{(2)} U^{(2)} \left(U^{(1)\intercal} F_{T}^{(1)\intercal} U^{(1)}\right)^\intercal X\right) X\right. \\
            \left.+ \left(\left(U^{(1)\intercal} U^{(2)\intercal} F_{T}^{(2)\intercal} U^{(2)} U^{(1)}\right)^\intercal X\right) X\right)
        \end{multlined} \\
        &= \!\begin{multlined}[t]
            F_{M}^{(3)} U^{(3)} F_{M}^{(2)} U^{(2)} F_{M}^{(1)} U^{(1)} X \\
            + \left(F_{M}^{(3)} U^{(3)} F_{M}^{(2)} U^{(2)} \left(U^{(1)\intercal} F_{T}^{(1)\intercal} U^{(1)}\right)^\intercal X\right) X \\
            + \left(F_{M}^{(3)} U^{(3)} \left(U^{(1)\intercal} U^{(2)\intercal} F_{T}^{(2)\intercal} U^{(2)} U^{(1)}\right)^\intercal X\right) X \\
            + \left(\left(U^{(3)\intercal} F_{T}^{(3)\intercal} U^{(3)}\right)^\intercal F_{M}^{(2)} U^{(2)} F_{M}^{(1)} U^{(1)} X\right) F_{M}^{(2)} U^{(2)} F_{M}^{(1)} U^{(1)} X
        \end{multlined} \\
        &= \!\begin{multlined}[t]
            F_{M}^{(3)} U^{(3)} F_{M}^{(2)} U^{(2)} F_{M}^{(1)} U^{(1)} X \\
            + \left(F_{M}^{(3)} U^{(3)} F_{M}^{(2)} U^{(2)} \left(U^{(1)\intercal} F_{T}^{(1)\intercal} U^{(1)}\right)^\intercal X\right) X \\
            + \left(F_{M}^{(3)} U^{(3)} \left(U^{(1)\intercal} U^{(2)\intercal} F_{T}^{(2)\intercal} U^{(2)} U^{(1)}\right)^\intercal X\right) X \\
            + \left(\left(U^{(1)\intercal} U^{(2)\intercal} U^{(3)\intercal} F_{T}^{(3)\intercal} U^{(3)} U^{(2)} U^{(1)}\right)^\intercal X\right) X
        \end{multlined} \\
        &=  \!\begin{multlined}[t]
            \left(F_M^{(3)} U^{(3)} F_M^{(2)} U^{(2)} F_M^{(1)} U^{(1)} +\right. \\
            \left.\left(\sum_{l=1}^{3} \left(F_M^{(3)} U^{(3)} F_M^{(2)} U^{(2)} \cdots F_M^{(l+1)} U^{(l+1)}\right) \left(U^{(1)\intercal} U^{(2)\intercal} \cdots U^{(l)\intercal} F_T^{(l)\intercal} U^{(l)} \cdots U^{(2)} U^{(1)}\right)^\intercal\right) X\right) X
        \end{multlined} \\
        &= \left(L_M^{(3)} + L_T^{(3)} X\right) X \text{,}
    \end{split}
\end{equation}
where, by induction:
\begin{equation}
    L_M^{(k)} = F_M^{(k)} U^{(k)} F_M^{(k-1)} U^{(k-1)} \cdots F_M^{(1)} U^{(1)}
\end{equation}
and
\begin{equation}
    L_T^{(k)} = \sum_{l=1}^k\left(F_M^{(k)} U^{(k)} F_M^{(k-1)} U^{(k-1)} \cdots  F_M^{(l+1)} U^{(l+1)}\right) \left(U^{(1)\intercal} U^{(2)\intercal} \cdots U^{(l)\intercal} F_T^{(l)\intercal} U^{(l)} \cdots U^{(2)} U^{(1)}\right)^\intercal \text{.}
\end{equation}
Here, ${}^\intercal$~denotes the transposition of the first two dimensions of tensors and matrix transposition.

\clearpage
\section{Selected Hyperparameter Values}
\label{sec:hyperparamter_values}

The hyperparameter values selected for each CNN and dataset are presented in Table~\ref{tab:hyperparameters}. These values provided higher accuracy during the validation step. The search space is presented in Table~\ref{tab:parameter_search_space}.

\begin{table}[!h]
    \caption{The hyperparameter values selected for each CNN and dataset using a Bayesian implemented by the Weights and Biases toolset (\texttt{https://www.wandb.com/}). For the \texttt{LeNet} and \texttt{LeNetCL} networks, the number of epochs was set to $200$ in training$^\dagger$, as the range of values shown in Table~\ref{tab:parameter_search_space} was proved insufficient to adjust the models.}
    \label{tab:hyperparameters}
    \centering
    \begin{tabular}{lccc}
        \toprule
        \texttt{BaseLinearNet} & MNIST & Fashion-MNIST & CIFAR-10 \\
        \midrule
        Batch size & 2868 & 3111 & 2198 \\
        Epochs & 44 & 15 & 41 \\
        Learning rate & 0.02039 & 0.05305 & 0.09827 \\
        Optimizer & Adam & Adam & Adam \\
        \toprule
        \texttt{BaseReLUNet} & MNIST & Fashion-MNIST & CIFAR-10 \\
        \midrule
        Batch size & 2838 & 3277 & 3677 \\
        Epochs & 50 & 48 & 21 \\
        Learning rate & 0.090331 & 0.08607 & 0.01 \\
        Optimizer & Adam & Adam & RMSprop \\
        \toprule
        \texttt{BaseReSProNet} & MNIST & Fashion-MNIST & CIFAR-10 \\
        \midrule
        Batch size & 2429 & 2339 & 3056\\
        Epochs & 50 & 45 & 19\\
        Learning rate & 0.7699 & 0.5601 & 0.4542 \\
        Optimizer & Adam & Adam & Adam \\
        \toprule
        \texttt{LeNet} & MNIST & Fashion-MNIST & CIFAR-10 \\
        \midrule
        Batch size & 2979 & 2444 & 2317 \\
        Epochs$^\dagger$ & 200 & 200 & 200 \\
        Learning rate & 0.004722 & 0.012 & 0.00159 \\
        Optimizer & RMSprop & Adam & RMSprop \\
        \toprule
        \texttt{LeNetCL} & MNIST & Fashion-MNIST & CIFAR-10 \\
        \midrule
        Batch size & 2088 & 2763 & 3595 \\
        Epochs$^\dagger$ & 200 & 200 & 200 \\
        Learning rate & 0.02 & 0.02302 & 0.01991\\
        Optimizer & Adam & Adam & Adam \\
        \bottomrule
    \end{tabular}
\end{table}

\begin{table}[!h]
    \caption{Search space for hyperparameters used in the experiments.}
    \label{tab:parameter_search_space}
    \centering
    \begin{tabular}{lccc}
        \toprule
        & Lower Bound & Upper Bound & Type \\
        \midrule
        Batch size & 2048 & 4096 & Discrete uniform distribution \\
        Epochs & 10 & 50 & Discrete uniform distribution \\
        Learning rate & 0.001 & 1.0 & Continuous uniform distribution \\
        \midrule
        Optimizer & \multicolumn{2}{c}{\{Adam, RMSprop\}} & Categorical distribution \\
        \bottomrule
    \end{tabular}
\end{table}

\clearpage
\section{Baseline Hyperparameter Sweep}
\label{sec:sweep_baseline}

\begin{figure}[!ht]
    \centering
    \begin{subfigure}[b]{0.65\textwidth}
        \centering
        \includegraphics[width=\textwidth]{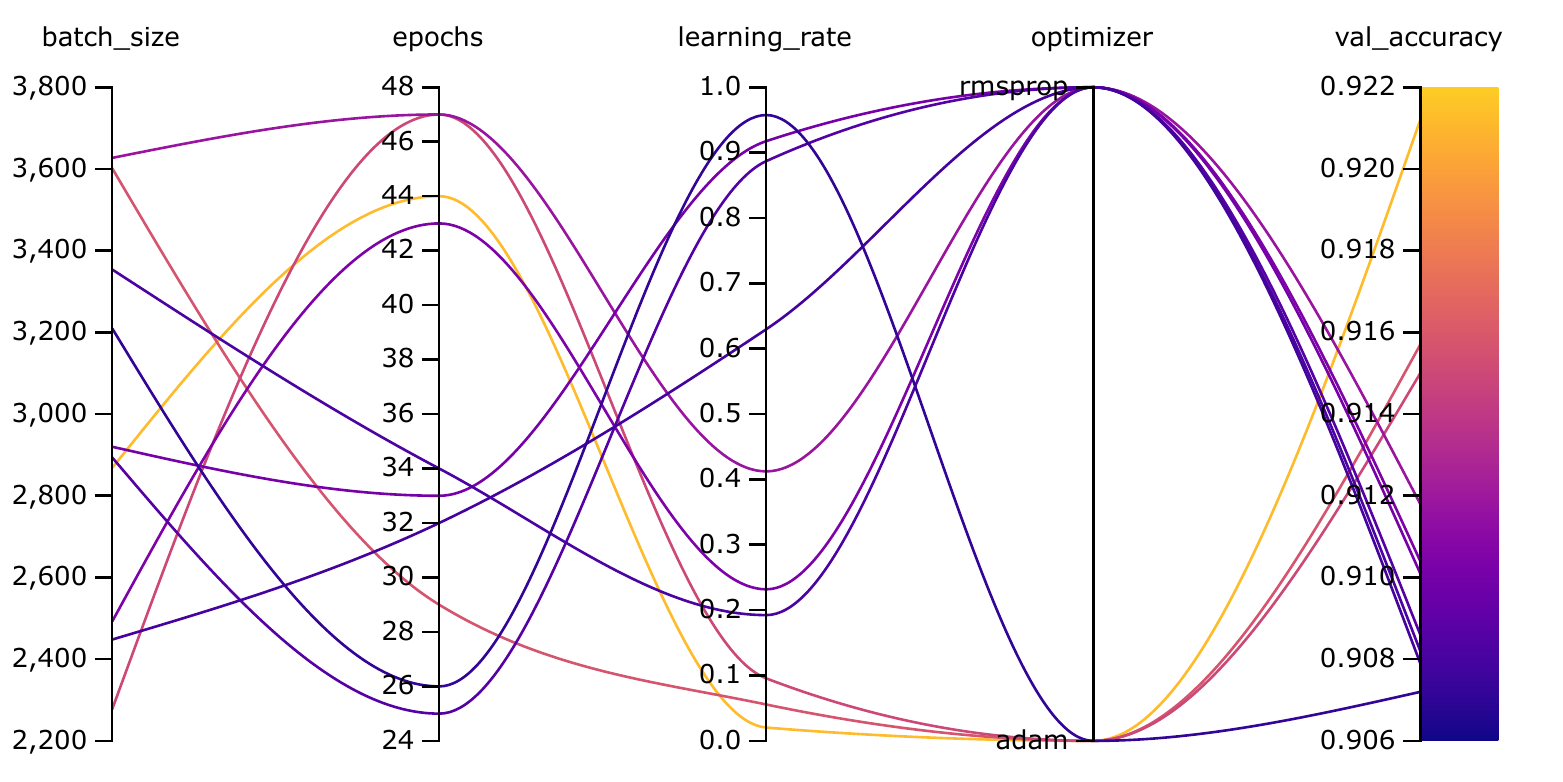}
        \caption{\texttt{BaseLinearNet}}
        \label{fig:BaseLinearNet_MNIST}
    \end{subfigure}
    \begin{subfigure}[b]{0.65\textwidth}
        \centering
        \includegraphics[width=\textwidth]{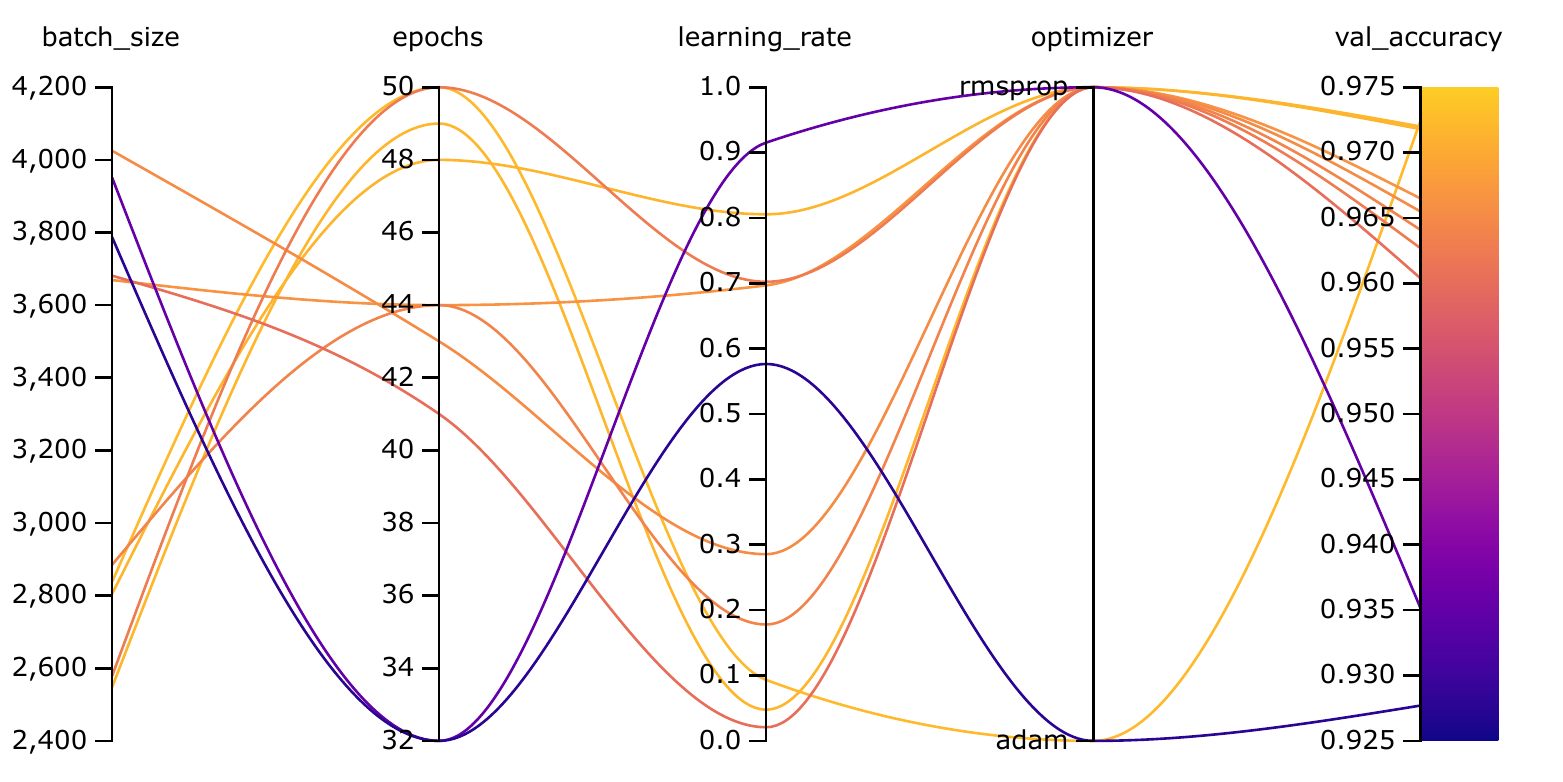}
        \caption{\texttt{BaseReLUNet}}
        \label{fig:BaseReLUNet_MNIST}
    \end{subfigure}
    \begin{subfigure}[b]{0.65\textwidth}
        \centering
        \includegraphics[width=\textwidth]{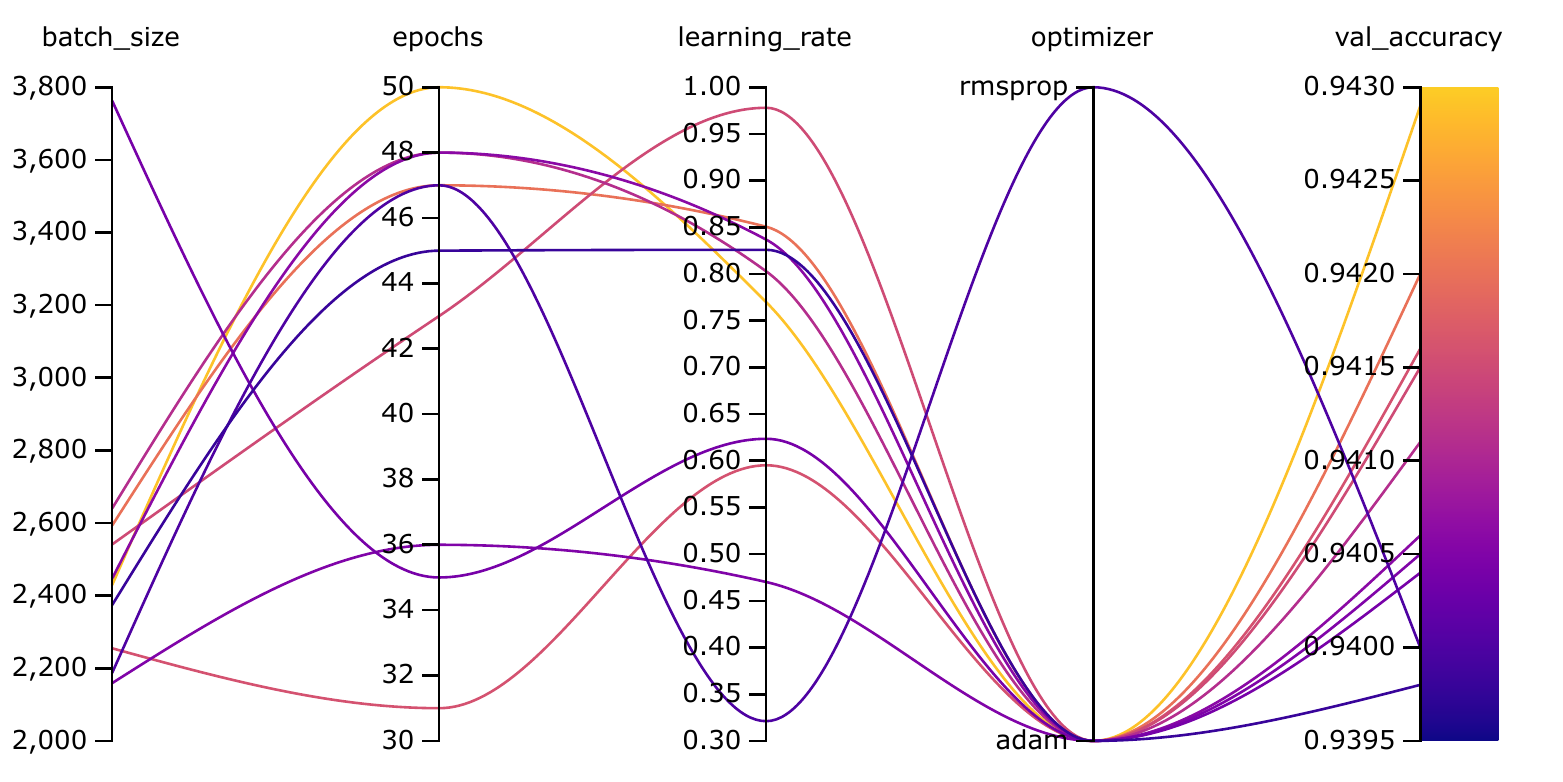}
        \caption{\texttt{BaseReSProNet}}
        \label{fig:BaseReSProNet_MNIST}
    \end{subfigure}
    \caption{Selected hyperparameters for the top-10 validation accuracy of CNNs used in Experiment~I on the MNIST dataset.}
    \label{fig:Baseline_MNIST}
\end{figure}

\begin{figure}[!ht]
    \centering
    \begin{subfigure}[b]{0.65\textwidth}
        \centering
        \includegraphics[width=\textwidth]{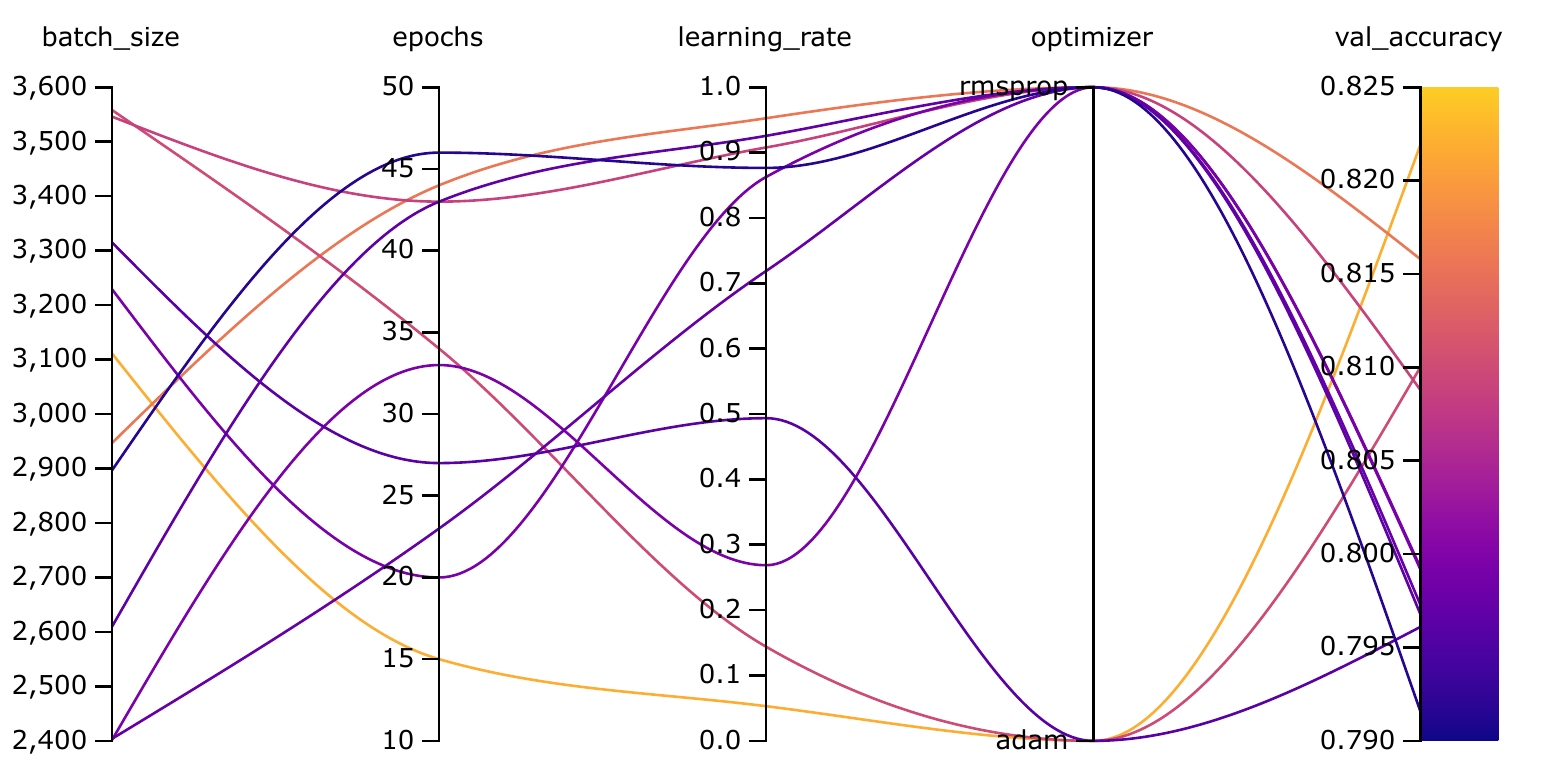}
        \caption{\texttt{BaseLinearNet}}
        \label{fig:BaseLinearNet_FashionMNIST}
    \end{subfigure}
    \begin{subfigure}[b]{0.65\textwidth}
        \centering
        \includegraphics[width=\textwidth]{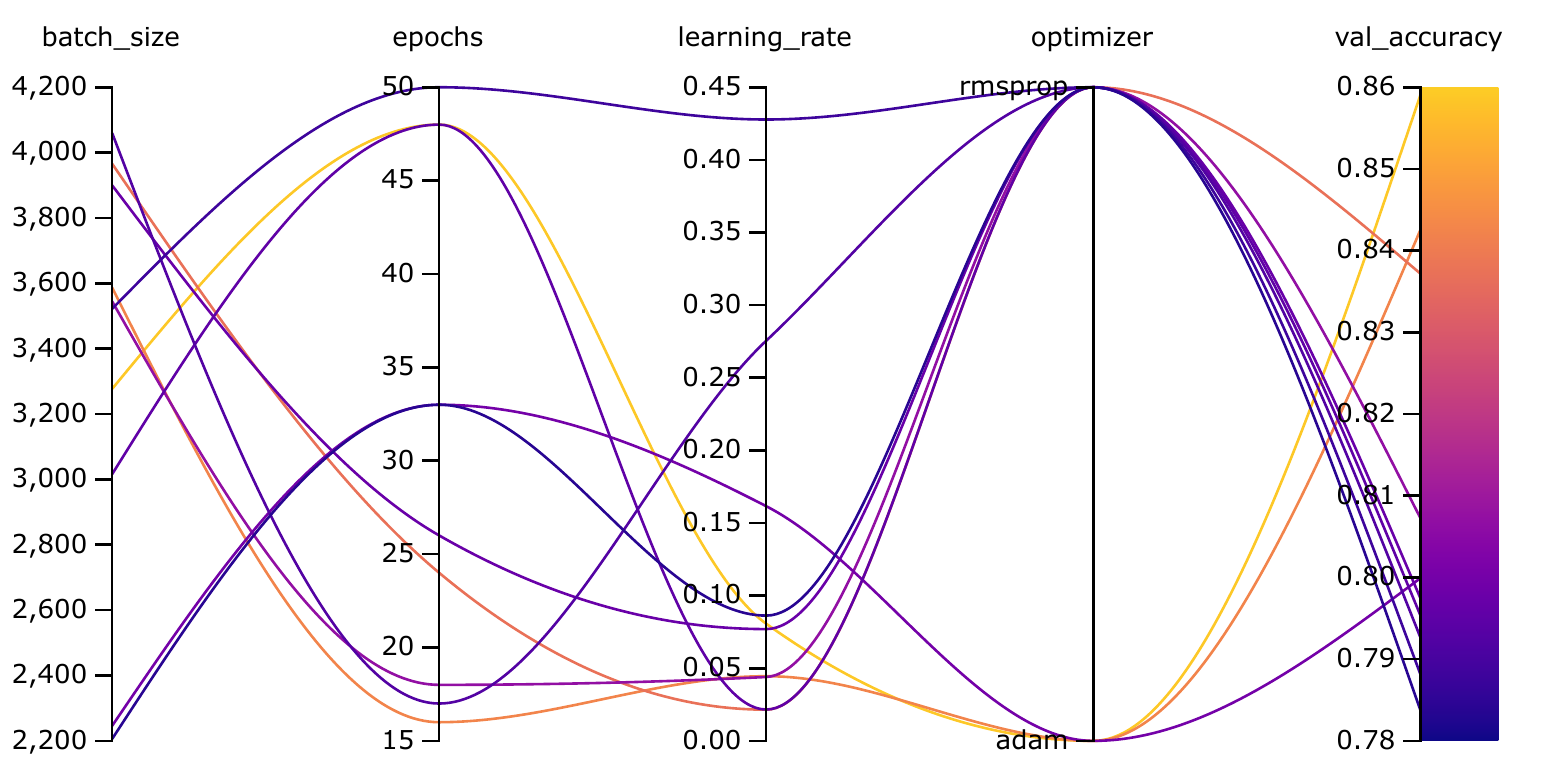}
        \caption{\texttt{BaseReLUNet}}
        \label{fig:BaseReLUNet_FashionMNIST}
    \end{subfigure}
    \begin{subfigure}[b]{0.65\textwidth}
        \centering
        \includegraphics[width=\textwidth]{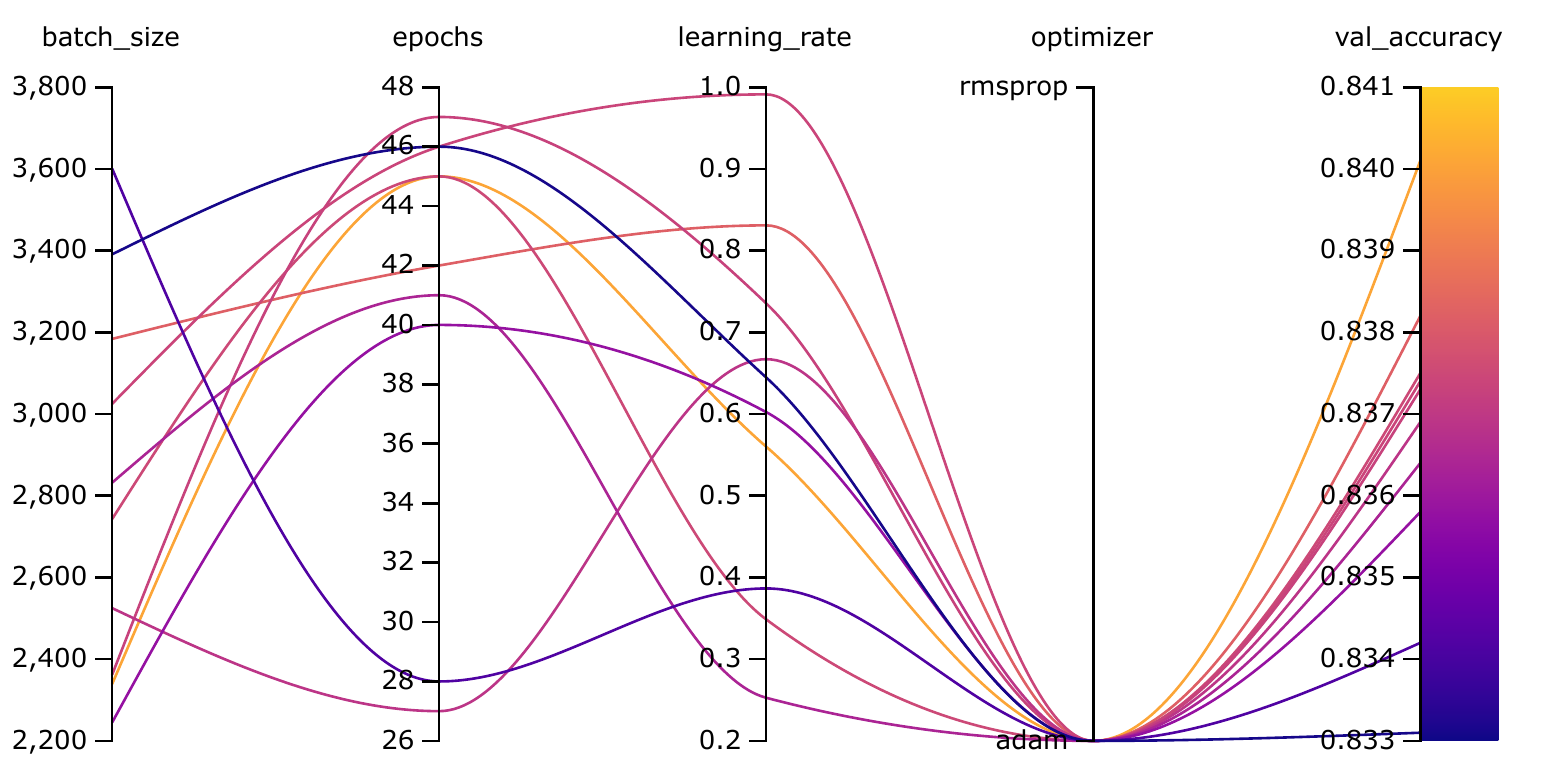}
        \caption{\texttt{BaseReSProNet}}
        \label{fig:BaseReSProNet_FashionMNIST}
    \end{subfigure}
    \caption{Selected hyperparameters for the top-10 validation accuracy of CNNs used in Experiment~I on the Fashion-MNIST.}
    \label{fig:Baseline_FashionMNIST}
\end{figure}

\begin{figure}[!ht]
    \centering
    \begin{subfigure}[b]{0.65\textwidth}
        \centering
        \includegraphics[width=\textwidth]{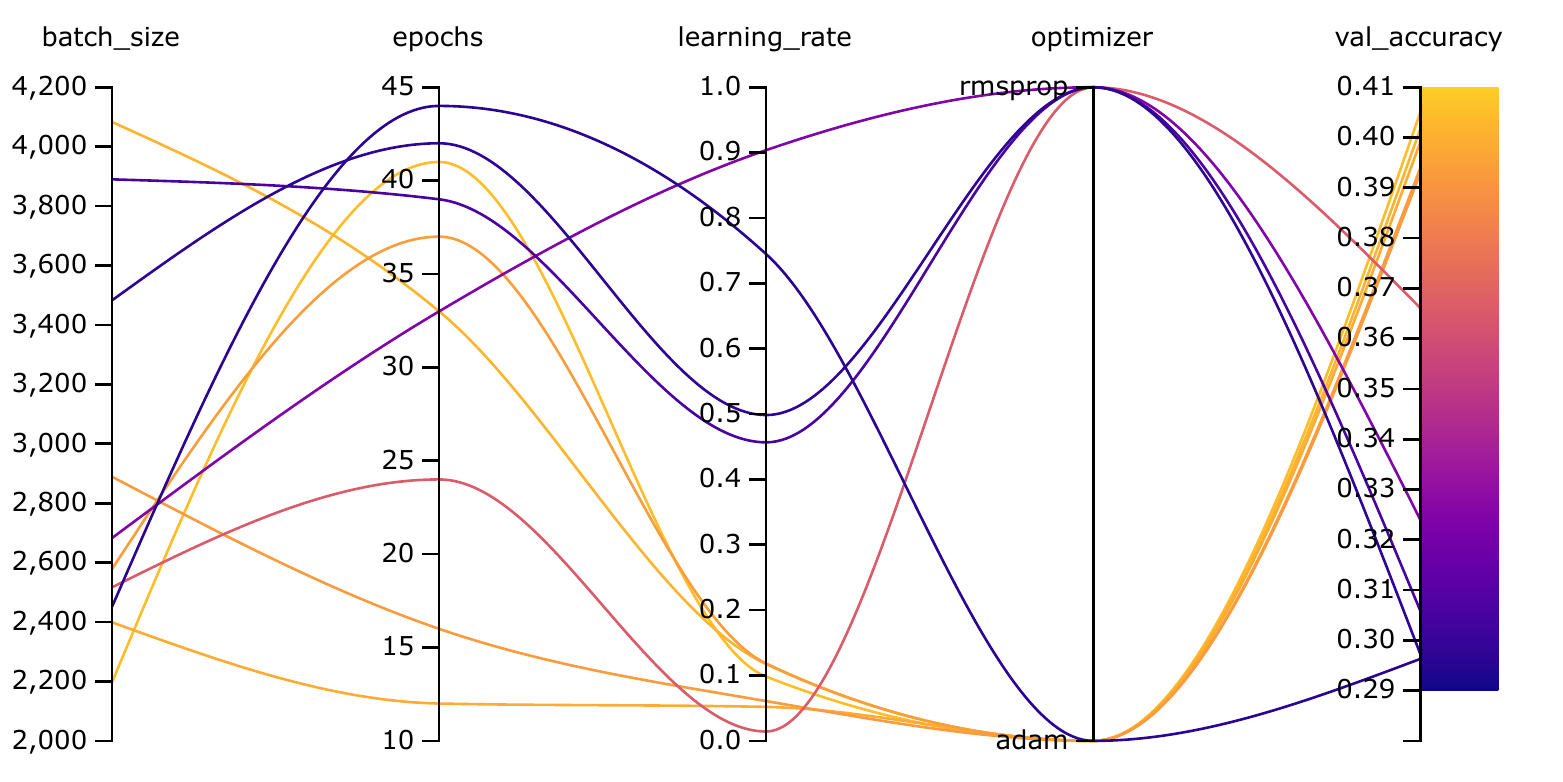}
        \caption{\texttt{BaseLinearNet}}
        \label{fig:BaseLinearNet_CIFAR10}
    \end{subfigure}
    \begin{subfigure}[b]{0.65\textwidth}
        \centering
        \includegraphics[width=\textwidth]{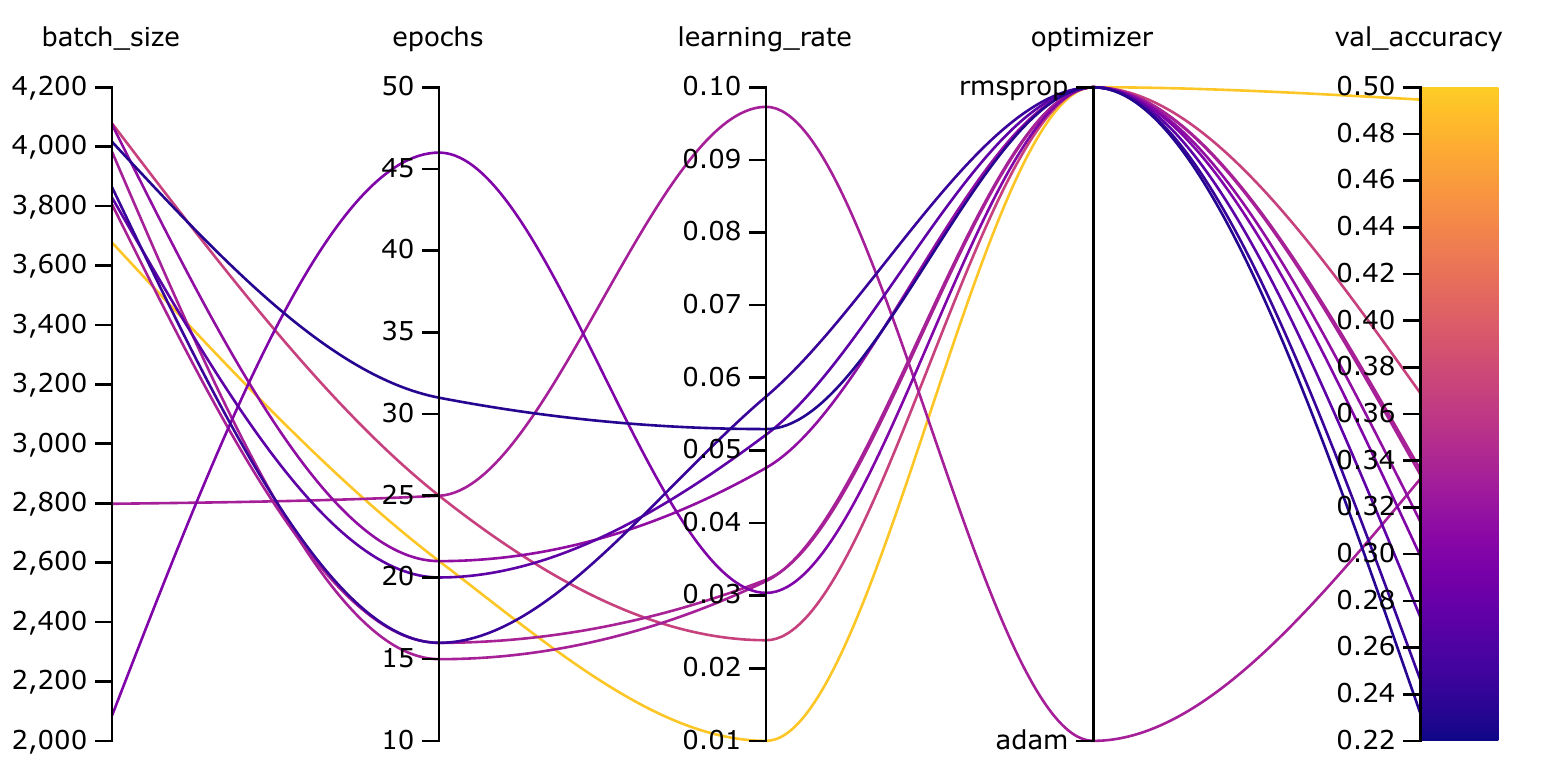}
        \caption{\texttt{BaseReLUNet}}
        \label{fig:BaseReLUNet_CIFAR10}
    \end{subfigure}
    \begin{subfigure}[b]{0.65\textwidth}
        \centering
        \includegraphics[width=\textwidth]{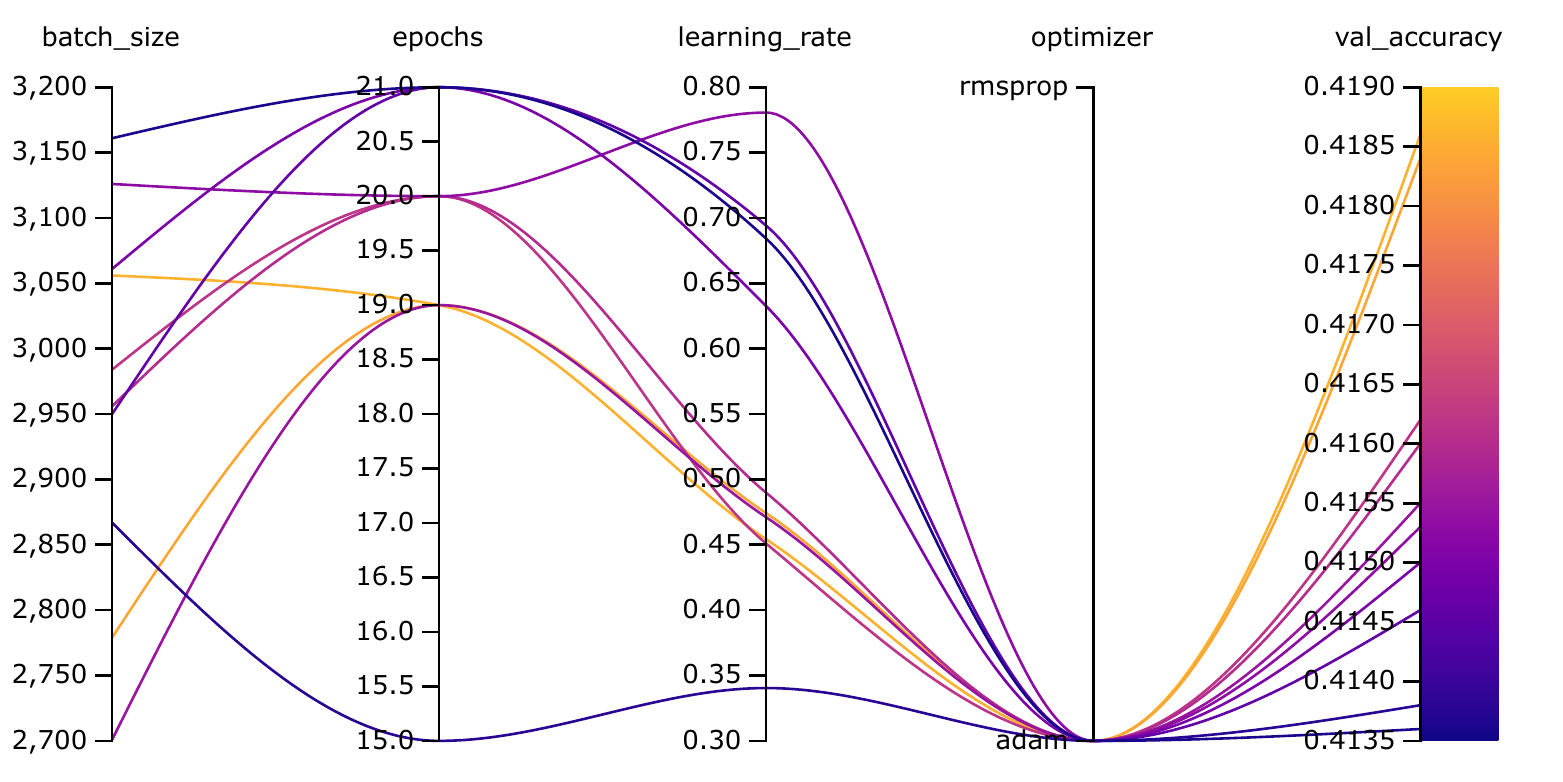}
        \caption{\texttt{BaseReSProNet}}
        \label{fig:BaseReSProNet_CIFAR10}
    \end{subfigure}
    \caption{Selected hyperparameters for the top-10 validation accuracy of CNNs used in Experiment~I on the CIFAR-10 dataset.}
    \label{fig:Baseline_CIFAR10}
\end{figure}

Figs.~\ref{fig:Baseline_MNIST}, \ref{fig:Baseline_FashionMNIST}, and~\ref{fig:Baseline_CIFAR10} present smooth parallel coordinate plots indicating the hyperparameter values selected for the top-10 validation accuracy achieved by \texttt{BaseLinearNet}, \texttt{BaseReLUNet}, and \texttt{BaseReSProNet} during training using, respectively, the MNIST~\cite{lecun2010mnist}, Fashion-MNIST~\cite{xiao2017fashion}, and CIFAR-10~\cite{Krizhevsky09learningmultiple} datasets.

\clearpage
\section{LeNet Hyperparameter Sweep}
\label{sec:sweep_lenet}

\begin{figure}[!ht]
    \centering
    \begin{subfigure}[b]{0.65\textwidth}
        \centering
        \includegraphics[width=\textwidth]{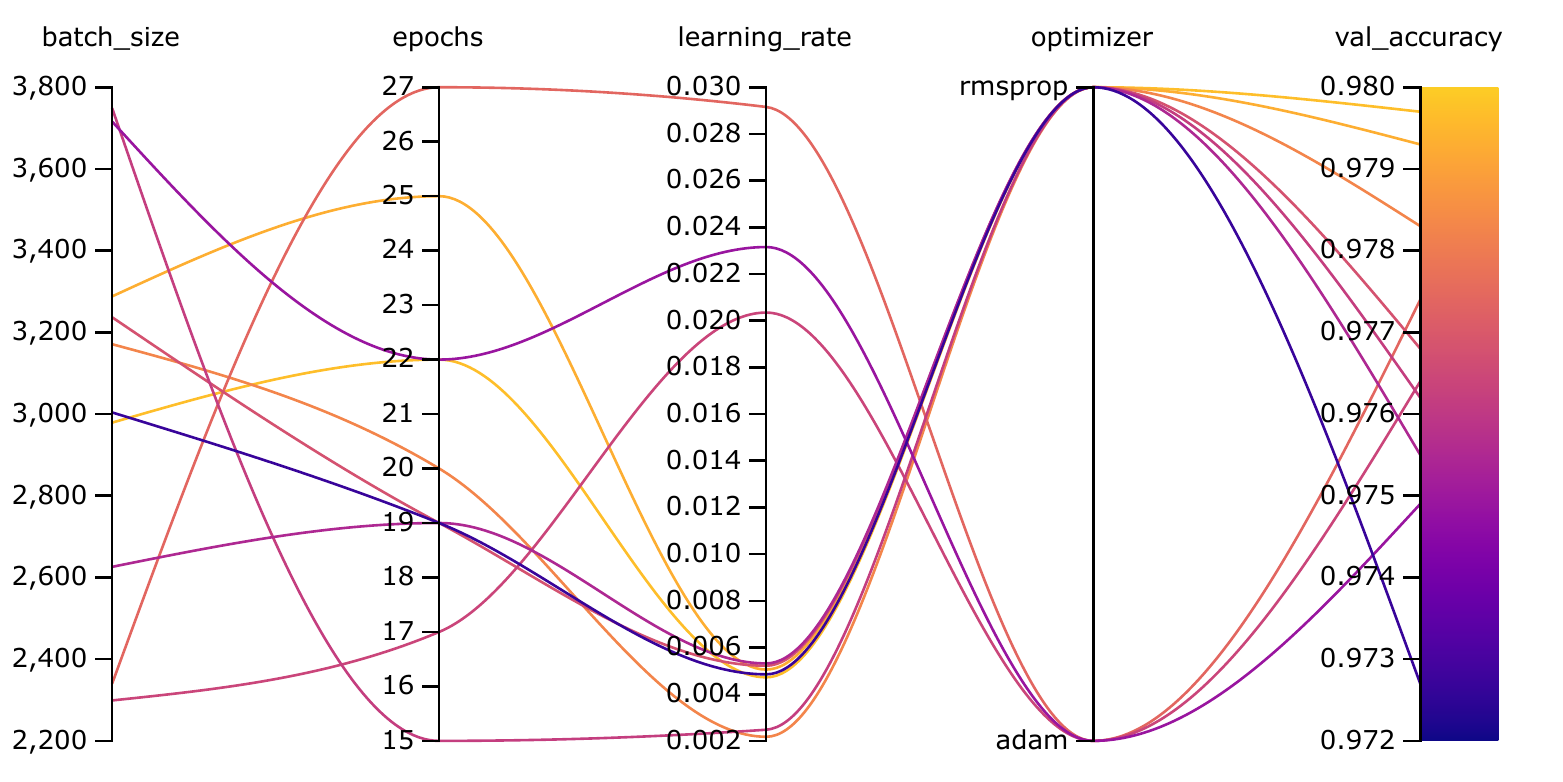}
        \caption{\texttt{LeNet}}
        \label{fig:LeNet_MNIST}
    \end{subfigure}
    \begin{subfigure}[b]{0.65\textwidth}
        \centering
        \includegraphics[width=\textwidth]{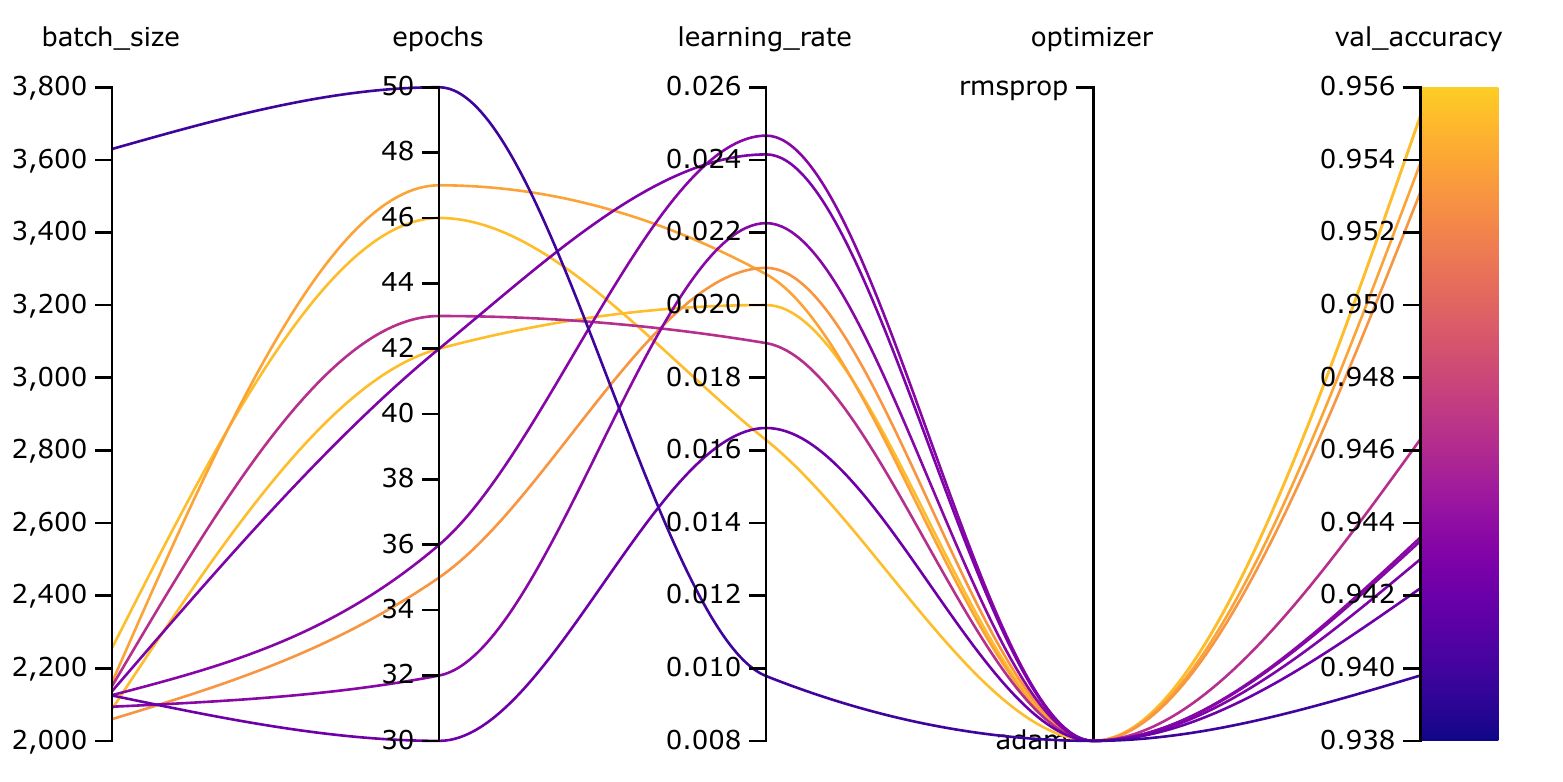}
        \caption{\texttt{LeNetCL}}
        \label{fig:LeNetCL_MNIST}
    \end{subfigure}
    \caption{Selected hyperparameters for the top-10 validation accuracy of CNNs based on the LeNet-5~\cite{lecun-01a}, used in Experiment~II on the MNIST dataset.}
    \label{fig:LeNetBased_MNIST}
\end{figure}

\begin{figure}[!ht]
    \centering
    \begin{subfigure}[b]{0.65\textwidth}
        \centering
        \includegraphics[width=\textwidth]{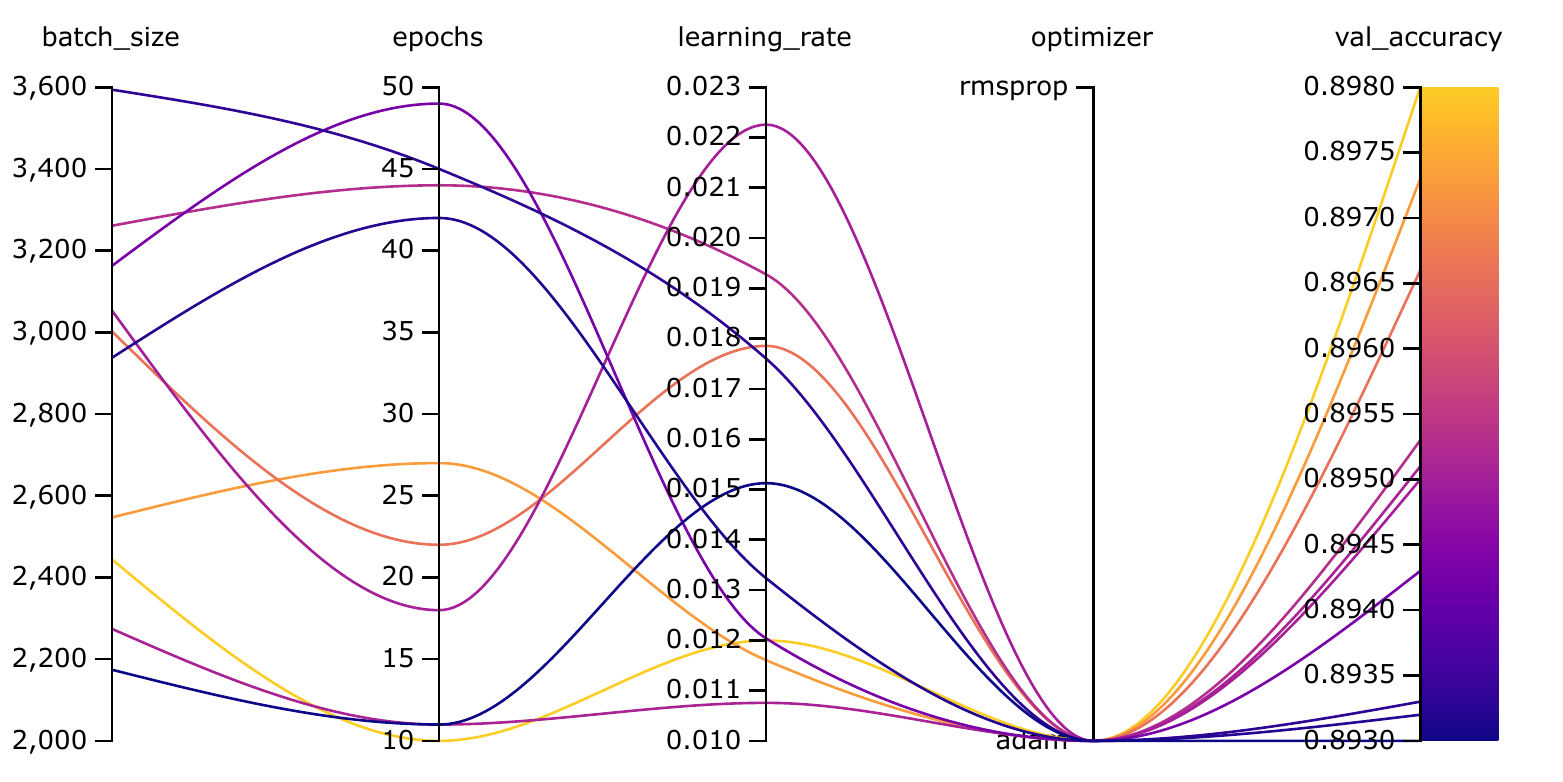}
        \caption{\texttt{LeNet}}
        \label{fig:LeNet_FashionMNIST}
    \end{subfigure}
    \begin{subfigure}[b]{0.65\textwidth}
        \centering
        \includegraphics[width=\textwidth]{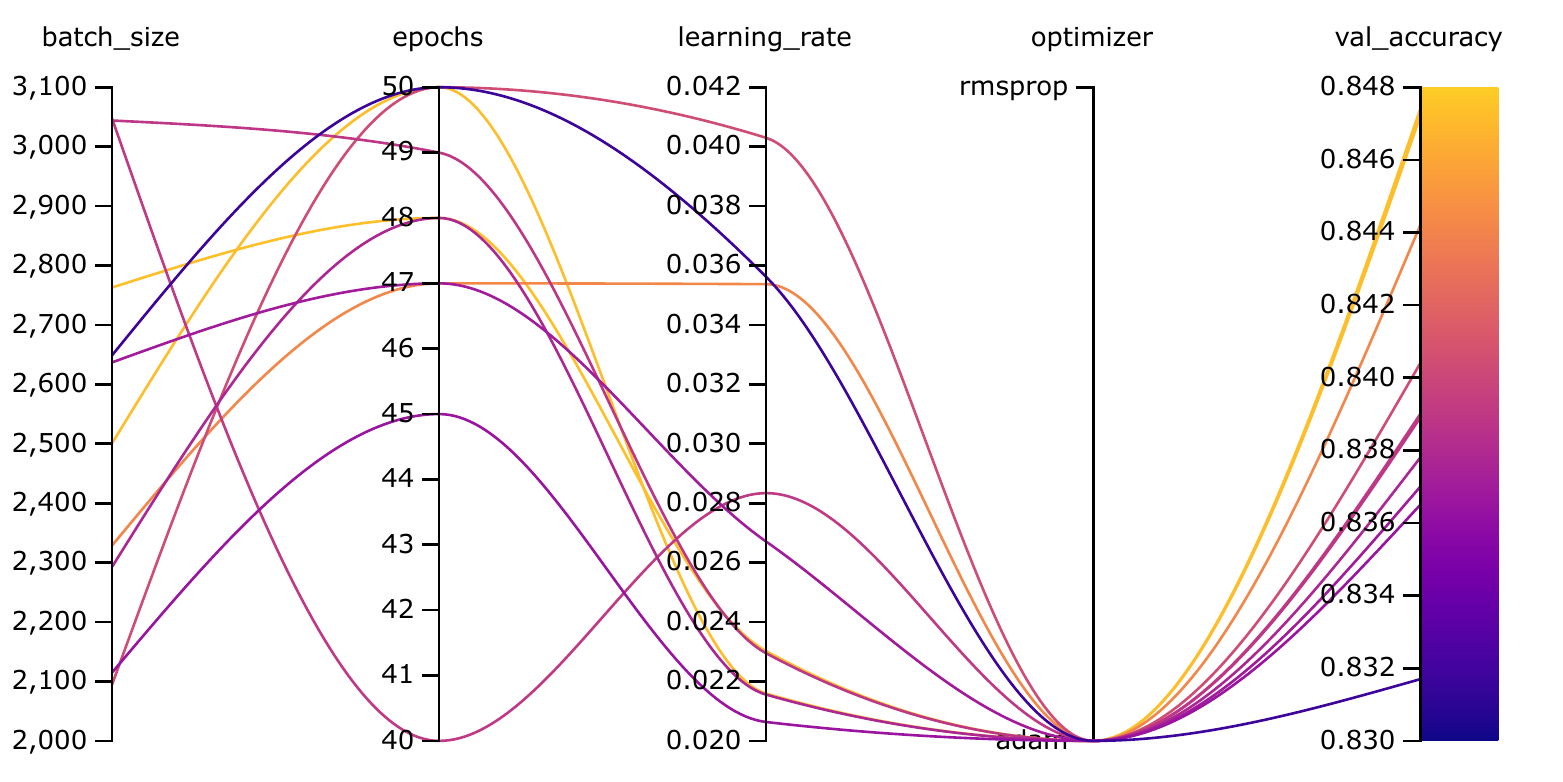}
        \caption{\texttt{LeNetCL}}
        \label{fig:LeNetCL_FashionMNIST}
    \end{subfigure}
    \caption{Selected hyperparameters for the top-10 validation accuracy of CNNs based on the LeNet-5~\cite{lecun-01a}, used in Experiment~II on the Fashion-MNIST dataset.}
    \label{fig:LeNetBased_FashionMNIST}
\end{figure}

\begin{figure}[!ht]
    \centering
    \begin{subfigure}[b]{0.65\textwidth}
        \centering
        \includegraphics[width=\textwidth]{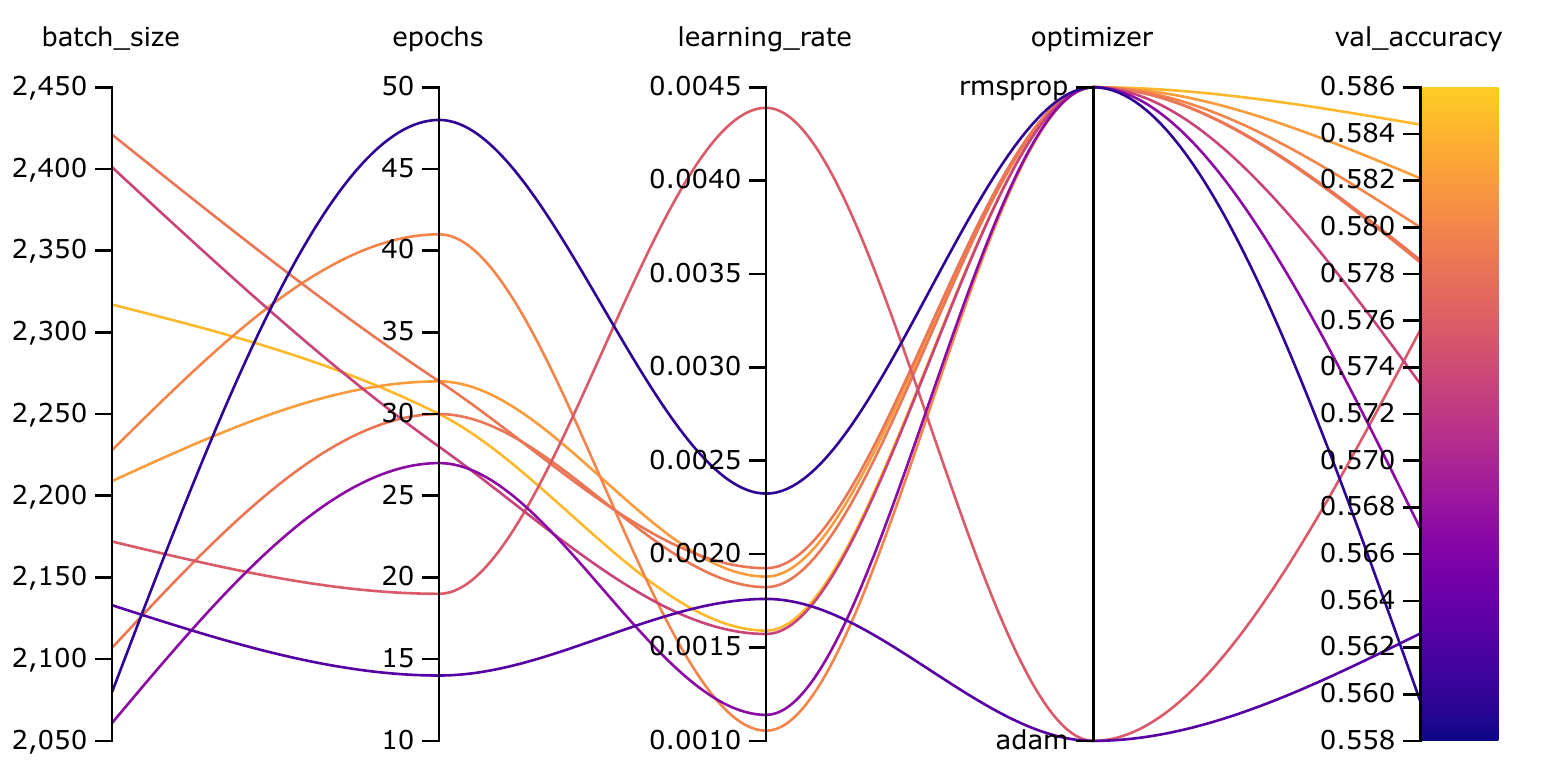}
        \caption{\texttt{LeNet}}
        \label{fig:LeNet_CIFAR10}
    \end{subfigure}
    \begin{subfigure}[b]{0.65\textwidth}
        \centering
        \includegraphics[width=\textwidth]{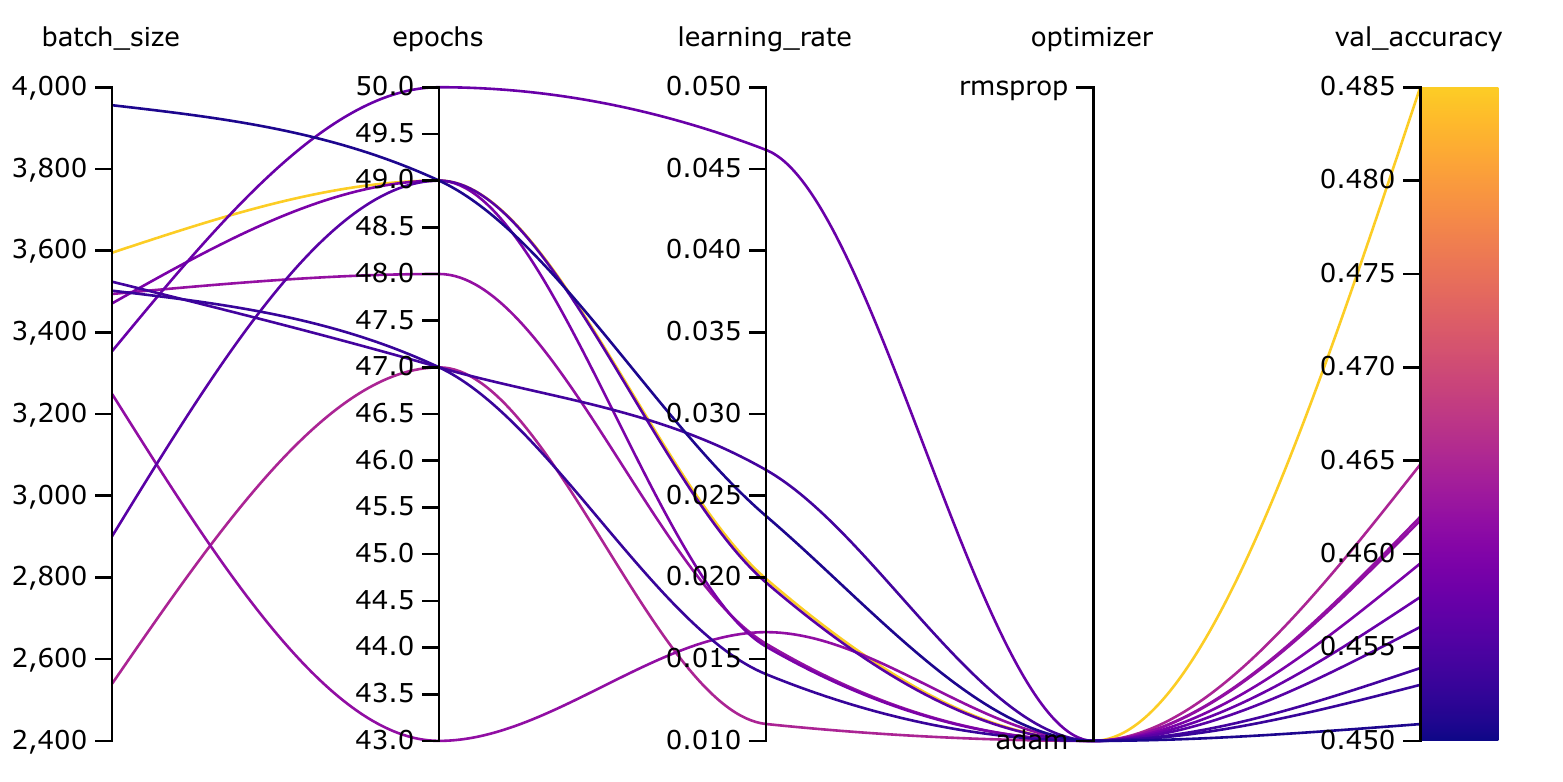}
        \caption{\texttt{LeNetCL}}
        \label{fig:LeNetCL_CIFAR10}
    \end{subfigure}
    \caption{Selected hyperparameters for the top-10 validation accuracy of CNNs based on the LeNet-5~\cite{lecun-01a}, used in Experiment~II on the CIFAR-10 dataset.}
    \label{fig:LeNetBased_CIFAR10}
\end{figure}

Figs.~\ref{fig:LeNetBased_MNIST}, \ref{fig:LeNetBased_FashionMNIST}, and~\ref{fig:LeNetBased_CIFAR10} present smooth parallel coordinate plots indicating the hyperparameter values selected for the top-10 validation accuracy achieved by \texttt{LeNet} and \texttt{LeNetCL} during training using, respectively, the MNIST~\cite{lecun2010mnist}, Fashion-MNIST~\cite{xiao2017fashion}, and CIFAR-10~\cite{Krizhevsky09learningmultiple} datasets. After this initial parameter selection, we have trained each CNN/dataset pair using $200$ epochs.

\bibliographystyle{IEEEtran}
\bibliography{references}